\documentclass{article}

\PassOptionsToPackage{numbers, compress}{natbib}

\usepackage[preprint]{paperstyle}
\usepackage{tabularx}
\usepackage{float}

\usepackage[utf8]{inputenc} %
\usepackage[T1]{fontenc}    %
\usepackage{hyperref}       %
\usepackage{url}            %
\usepackage{booktabs}       %
\usepackage{amsfonts}       %
\usepackage{array}
\usepackage{nicefrac}       %
\usepackage{microtype}      %
\usepackage{xcolor}         %
\usepackage{amsmath}
\usepackage[numbers]{natbib}
\usepackage{booktabs}
\usepackage{multirow}
\usepackage{graphicx}
\usepackage{algorithm}
\usepackage{bm}
\usepackage{algpseudocode}
\usepackage{wrapfig}
\usepackage{threeparttable}

\usepackage{caption}
\usepackage{tikz}
\usetikzlibrary{bayesnet}

\usepackage{xcolor}

\newcommand{\MoralMachines}{\textsc{Moral Machine}}
\newcommand{\Jokes}{\textsc{Jokes}}
\newcommand{\Population}{\textsc{Population}}
\newcommand{\HalfLives}{\textsc{Half-lives}}

\newcommand{\GemmaFour}{Gemma~4 31B}
\newcommand{\GemmaThree}{Gemma~3 12B}
\newcommand{\DeepSeek}{DeepSeek-R1 14B}
\newcommand{\GptOss}{GPT-OSS 20B}
\newcommand{\OLMo}{OLMo~3 7B}
\newcommand{\QwenSmall}{Qwen~2.5 14B}
\newcommand{\QwenLarge}{Qwen~3.6 35B}

\newcommand{\keff}{k_{\mathrm{eff}}}

\newcommand{\MBT}{MBT}

\newcommand{\DeltaMSE}{\Delta\mathrm{MSE}}

\title{Multiple latent orderings better predict language model preferences}

\author{%
  Aviral Chawla\thanks{Equal contribution.} \quad
  William H.W. Thompson\footnotemark[1] \\
  Vermont Complex Systems Institute \\
  University of Vermont \\
  Burlington, Vermont 05403 \\
  \texttt{\{achawla1,wthomps3\}@uvm.edu}
  \And
  Jean-Gabriel Young \\
  Vermont Complex Systems Institute \\
  Department of Statistics \\
  University of Vermont \\
  Burlington, Vermont 05403 \\
  \texttt{jyoung22@uvm.edu}
}

\begin{document}

\maketitle

\begin{abstract}
    Language models are frequently employed in settings where they are asked to make value judgments and choices. 
    These observed choices often exhibit intransitivity: A model may prefer item $A$ to $B$ and $B$ to $C$, while also preferring $C$ to $A$.
    Existing work that models LLM preferences treats such inconsistencies as sampling noise around a single latent ordering. 
    We instead propose that intransitivity reflects the aggregation of multiple latent, internally consistent orderings.
    We first show that observed inconsistencies cannot be explained by a single ordering under any monotone link function.
    We then introduce a noise-augmented mixture Bradley--Terry (MBT) model that infers latent preference components from repeated pairwise comparisons. 
    Across seven models and four tasks, a mixture of orderings often explains structural inconsistencies better than single-utility models. 
    We find that aggregate preferences often hide underlying preference heterogeneity. 
    A case study on Moral Machine dilemmas shows that models which disagree on aggregate orderings can still share latent components. 
    Together, these results suggest that LLMs reflect plural preferences. 
    Alignment and evaluation pipelines that treat LLM preferences as a single function, therefore, risk averaging over coherent orderings that different users may endorse differently.
\newline

\end{abstract}

\begin{figure}
    \centering
    \includegraphics[width=\linewidth]{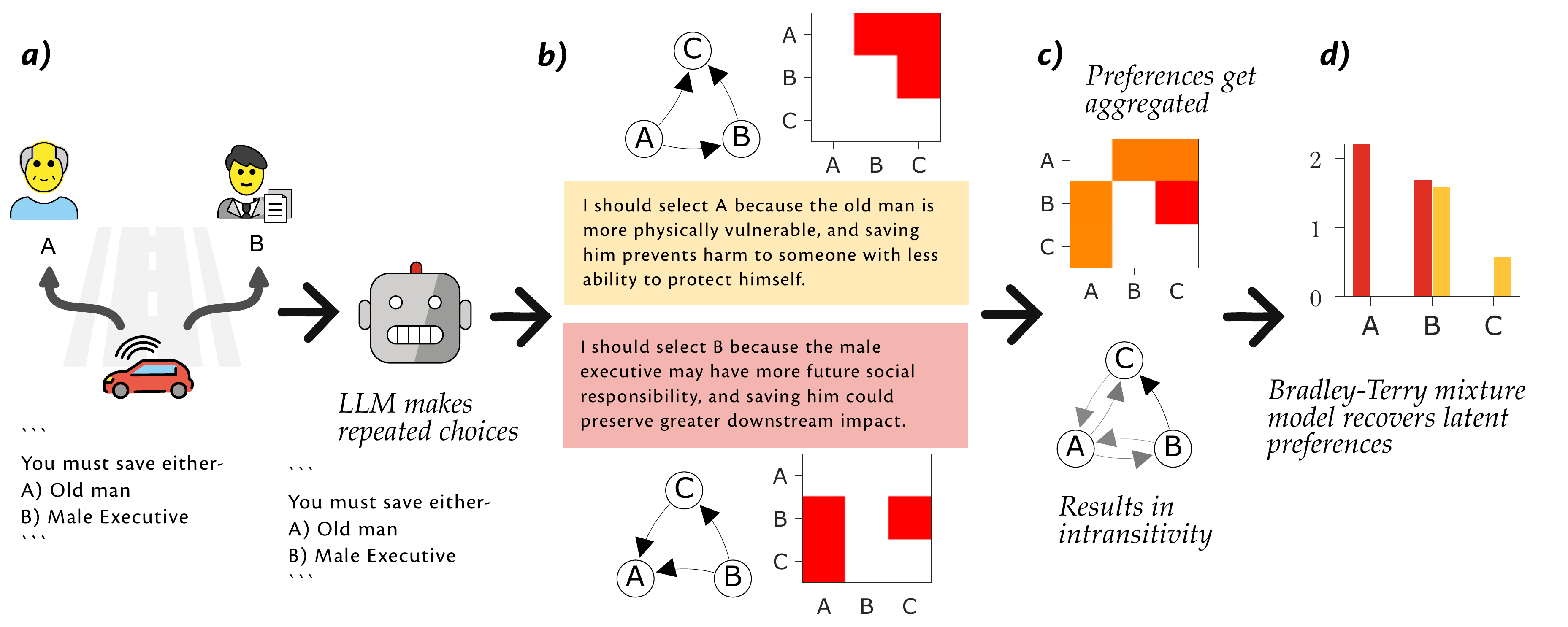}
    \caption{\textbf{Conceptual overview.} \textbf{a)} We elicit forced pairwise preferences from an LLM over a fixed set of items. \textbf{b)} The model's choices are shaped by competing preferences. For example, under some set of preferences, the model chooses $A\succ B$ ($A$ over $B$), $A\succ C$ and $B\succ C$ (top), while it prefers $B\succ A$, $B\succ C$, $C\succ A$ under a different set of preferences (bottom). \textbf{c)} Aggregating these samples yields intransitive cycles that look incoherent. \textbf{d)} However, the \MBT\ model allows us to recover both underlying preference modes from the aggregated output.}
    \label{fig:main}
\end{figure}

\section{Introduction}

Practitioners routinely use language models in ways that require them to make subjective judgments and decisions: simulating survey respondents, assisting with peer review, or answering ethical questions, typically in benchmarking studies~\cite{aher_using_nodate, dominguez-olmedo_questioning_nodate, wang_large_2025, autenrieth_how_2026}. 
The resulting choices can be thought of as evidence about what a model prefers, values, or would recommend.

This makes model preferences themselves an important area of evaluation given the scale and sensitivity of these tasks.
Prior work has explored LLM preferences in domains where the stakes are social, ethical, or institutional~\cite{aher_using_nodate, dominguez-olmedo_questioning_nodate, wang_large_2025, autenrieth_how_2026, mazeika_utility_2025}. 
The standard output of such analyses is usually an aggregate preference: a single ranking, score, or reward-like ordering.

However, a large literature shows that LLM judgments are not always internally consistent~\cite{zheng_judging_2023, goli_can_nodate, xu_investigating_2025, wang_trustjudge_2025, moore_are_2024, nuenen_fragility_2026}. 
Repeated comparisons can produce \textit{intransitive cycles}, whereby a model prefers $A$ to $B$, $B$ to $C$, but also $C$ to $A$. 
These failures are typically treated as noise around a single latent ordering, which can be recovered through Bradley--Terry-style aggregation~\cite{mazeika_utility_2025, chiang_chatbot_2024, xu_judge-aware_2026, qian_who_2026, zeng_llm-rankfusion_2024, yu_elspr_2025}. 

But intransitivity need not mean arbitrary noise. 
Social choice theory has long shown that aggregating individually transitive preferences can produce collective cycles~\cite{Condorcet1785, may_intransitivity_1954, hagele_llulls_2001}.
Reinforcement learning acknowledges the importance of disaggregating collective human preference data when building reward models ~\cite{chen_pal_2024, shirali_direct_nodate}. 
Analogously, aggregating over multiple coherent orderings may explain the apparent inconsistency in LLM preferences.\footnote{Throughout when we refer to LLM preferences, or internal preferences, we mean that phenomenologically LLMs behavior can be explained \emph{as if} it were generated from preference orderings.
We do not imply any mechanistic explanation}

This matters because aggregate evaluation does not match how users experience LLM systems.
A user only receives one sampled response.
But if this draw can in fact come from different latent preference modes, then the aggregate can make a model's apparent preference look stable while individual interactions remain meaningfully different.

\subsection{Contribution}

We repeatedly prompt seven open-weight LLMs on four pairwise comparison tasks. Two are factual numerical tasks of varying difficulty: comparisons of US city populations (easy) and of atomic half-lives (hard). The other two are subjective: aesthetic comparisons of jokes, and moral comparisons drawn from the Moral Machine dilemmas \cite{awad2018moral}, where a runaway car must choose between two groups of individuals with different traits. We find:

\textbf{Structured noise.} Under a non-parametric diagnostic, no single preference utility, however noisy, can explain the intransitivities we see in LLM preferences. These results still hold when we control for prompt, order, and label biases.

\textbf{Predictive gains.} Our \MBT\ model predicts held-out LLM preference data better than a single-component Bradley--Terry baseline.

\textbf{Misleading aggregations.} Distinct components of two models that are aligned in aggregate may disagree with each other when disaggregated and vice-versa. 
Aggregation hides similarities and differences between models. 

Our results suggest that LLMs may not in fact be singular but plural. Our methods provide a framework to identify and manage that plurality, in the same way pluralistic alignment in RLHF identifies and manages plurality across human groups.

\section{Related Work}

\paragraph{LLMs as simulated respondents.}
A growing literature uses LLMs as simulated respondents in
social-science experiments and surveys, including Aher et al.'s
``Turing experiments''~\citep{aher_using_nodate} and
demographic-conditioning frameworks for human-sample
simulation~\citep{Argyle2023}. This is directly relevant:
pairwise preference elicitation treats the model as a respondent
whose choices reveal a value or judgment structure. But LLM
responses need not faithfully represent any human population.
Dominguez-Olmedo et al.~\citep{dominguez-olmedo_questioning_nodate}
document systematic distortions against American Community Survey
statistics, and synthetic respondents can flatten or misrepresent
identity groups~\citep{wang_large_2025, Bisbee2024}. We
extend this critique inward: rather than ask whether the model
matches a human population in aggregate, we ask whether a single
model's repeated choices aggregate over multiple latent orderings
of its own.

\paragraph{LLM preferences are unstable and intransitive.}
Prior work documents that LLM judgments are sensitive to
elicitation protocol, option order, baseline choice, and surface
form~\citep{zheng_judging_2023, pezeskhpour_large_2024,
wang_large_2025, zhou_fairer_2024}. Most
directly, Xu et al.~\citep{xu_investigating_2025} show that LLM
judges produce non-transitive pairwise preferences that make
rankings depend on the baseline; related work identifies cycles in
LLM rankings and proposes methods to recover stable global
leaderboards~\citep{zeng_llm-rankfusion_2024, yu_elspr_2025}. These
papers establish the phenomenon but treat intransitivity as a
reliability problem to be corrected. We instead propose it is a signal of model multiplicity.

\paragraph{Plural preferences in humans and LLMs.}
Pluralistic alignment argues that human preferences are
heterogeneous and should not be collapsed into a single reward or
ranking. Recent methods capture this through distributional
rewards, latent annotator types, user-specific variables, and
mixture models~\citep{siththaranjan_distributional_2024,
chidambaram_direct_2025, poddar_personalizing_2024, chen_pal_2024,
shen_micro_2025, shirali_direct_nodate}. A related literature
measures values, moral judgments, and personalities expressed by
LLMs~\citep{mazeika_utility_2025, moore_are_2024,
nuenen_fragility_2026, song_identifying_2024}; Song et
al.~\citep{song_identifying_2024} in particular argue that LLMs
can exhibit multiple personality profiles, close in spirit to our
claim of multiple latent modes within a single model. Our work
shifts the target: pluralistic alignment locates heterogeneity in
the human population outside the model, and value-measurement
reports aggregate traits. We instead ask whether heterogeneity
appears inside a fixed model's repeated choices, inferring latent
orderings from pairwise comparisons rather than from open-ended text or survey batteries.

\section{Background}

\textbf{Pairwise preferences from language models}\label{sec:pairwise_pref_llm}
We elicit pairwise preferences from a range of open-weight language models on four datasets, using domain-relevant questions that compare their items (see Table~\ref{tab:datasets}). 
Two datasets are in the factual domains, \Population and \HalfLives~\cite{wolfram_world_nodate}, and the other two invite more subjective preferences, \Jokes~\cite{simpson_predicting_2019} and \MoralMachines~\cite{awad2018moral}.
The factual domains differ in difficulty: U.S. city populations provide a relatively familiar numerical comparison task, while atomic half-lives offer a harder scientific comparison task~\cite{wolfram_world_nodate}.

\begin{table}[h]
    \centering
    \caption{Datasets used in our pairwise preference experiments.}
    \label{tab:datasets}
    \small
    \setlength{\tabcolsep}{4pt}
    \begin{tabularx}{\linewidth}{@{}l >{\raggedright\arraybackslash}X r >{\raggedright\arraybackslash}X@{}}
    \toprule
    Data & Description & Items & Comparison question \\
    \midrule
    \Jokes &
    Pairwise humor judgments from Simpson et al.~\cite{simpson_predicting_2019}. &
    50 &
    Which joke is funnier? \\
    \addlinespace
    \MoralMachines &
    Autonomous-vehicle moral dilemmas from the Moral Machine experiment~\cite{awad2018moral}. &
    50 &
    Which person should be sacrificed? \\
    \addlinespace
    \Population &
    U.S. Census population data for U.S. cities. &
    50 &
    Which city has the greater population? \\
    \addlinespace
    \HalfLives &
    Atomic half-lives retrieved from the IAEA LiveChart of Nuclides API~\cite{iaea_livechart}. &
    50 &
    Which isotope has the longer half-life? \\
    \bottomrule
    \end{tabularx}
\end{table}

Each LLM sees every unordered item pair under four prompt variants and both presentation orders, with $M$ samples per condition ($8M$ attempted comparisons per pair). Human comparison counts vary by pair because they come from existing crowdsourced datasets.

In \MoralMachines\ trials, the model selects the person to sacrifice. We store the selected person as the pairwise ``winner'' for likelihood construction. Higher Bradley--Terry scores therefore mean a person is more likely to be sacrificed; lower scores indicate greater protection.

For each combination of dataset and model, we force a binary choice over every possible pair of items.
The items are given to the model with randomly assigned pseudo-identifiers, and the model is instructed to return structured output indicating the selected identifier.
Informed by prior literature, the generation protocol controls for three potential sources of bias: presentation order~\cite{pezeshkpour2024large, wang_fair_2024}, prompt wording~\cite{zhou_fairer_2024}, and option labels~\cite{zheng2023large}.
In a nutshell, each unordered pair of items is presented to the models in both possible orders, with various prompt rewordings that preserve semantics and randomized pseudo-identifiers; Appendix~\ref{app:preference-protocol} gives details.

\textbf{Choice theory} models the preferences of an agent over a set of items $[N] = \{1, \ldots, N\}$ with a preference relation $\succ$, where $i \succ j$ means item $i$ is preferred to item $j$.
Theories usually start with two axioms; \emph{completeness} requires that for any pair $i, j$, either $i \succ j$, $j \succ i$, or $i \sim j$ when the chooser is indifferent;
\emph{transitivity} means that $i \succ j$ and $j \succ k$ imply $i \succ k$.
A set of relations that satisfy both is a \emph{total order}.
This means that there is a ranking such that $i_1 \succ i_2 \succ \cdots \succ i_N$. 
Any total ordering of a finite set of items also implies that there is an ordinal utility function $U : [N] \to \mathbb{R}$ with $U(i) > U(j) \iff i \succ j$. A more comprehensive review of social choice theory can be found in \cite{brandt_introduction_2016}.

\section{Preference inconsistencies are structured}
\label{sec:structured-inconsistency}

Existing literature operationally assumes that intransitivities in LLM choices are sampling noise around a single latent ranking. Bradley--Terry models underlie reward modeling in RLHF~\citep{Christiano2017,Ouyang2022} as well as LLM-as-judge evaluations~\citep{zheng_judging_2023}. Most directly, Mazeika et al.~\citep{mazeika_utility_2025} fit a Thurstonian model to LLM pairwise preferences. All of these assume that each item $i$ has a latent score $\theta_i$, and that the LLM's choice between $i$ and $j$ is determined by drawing independent noise terms
$\epsilon_i, \epsilon_j$ and selecting $i$ if
$\theta_i + \epsilon_i > \theta_j + \epsilon_j$. Models in this family, where choices arise from comparing latent scores plus i.i.d. noise, are called \textit{random utility models}; under any such model, the underlying preferences are coherent and apparent intransitivities are
an artifact of the noise. Our alternative hypothesis is that
inconsistencies in LLM preferences are not solely due to sampling noise, and we test this assumption directly.

\textbf{Null model.} We compare observed preference patterns against a generic random utility null model. All random utility models, including Thurstone
and Bradley--Terry, satisfy \emph{strong stochastic transitivity} (SST):
$p_{ik} \geq \max(p_{ij}, p_{jk})$ whenever $p_{ij}, p_{jk} \geq 1/2$.
The mean SST deficit
\begin{equation}
D^{\text{SST}} = \frac{1}{|T|}\sum_{(i,j,k) \in T}
\max\bigl(\max(p_{ij}, p_{jk}) - p_{ik},\; 0\bigr)
\end{equation}
measures the average positive violation across Borda-ordered triplets
$T = \{(i,j,k) : b_i \geq b_j \geq b_k\}$, where $b_i$ is the Borda
count for item~$i$. The \emph{Borda count} $b_i = \frac{1}{N-1}\sum_{j \neq i} \hat{p}_{ij}$ is the average empirical win probability of item~$i$ across all comparisons. We adapt work by \citet{Singh2025} to build a non-parametric null that asks how large $D^{\text{SST}}$ should be under \emph{any} random utility model, and compare the observed deficit against bootstrap samples from this null. Full construction and inference details appear in
Appendix~\ref{app:null-model}.

\begin{figure}[t]
    \centering
    \includegraphics[width=\linewidth]{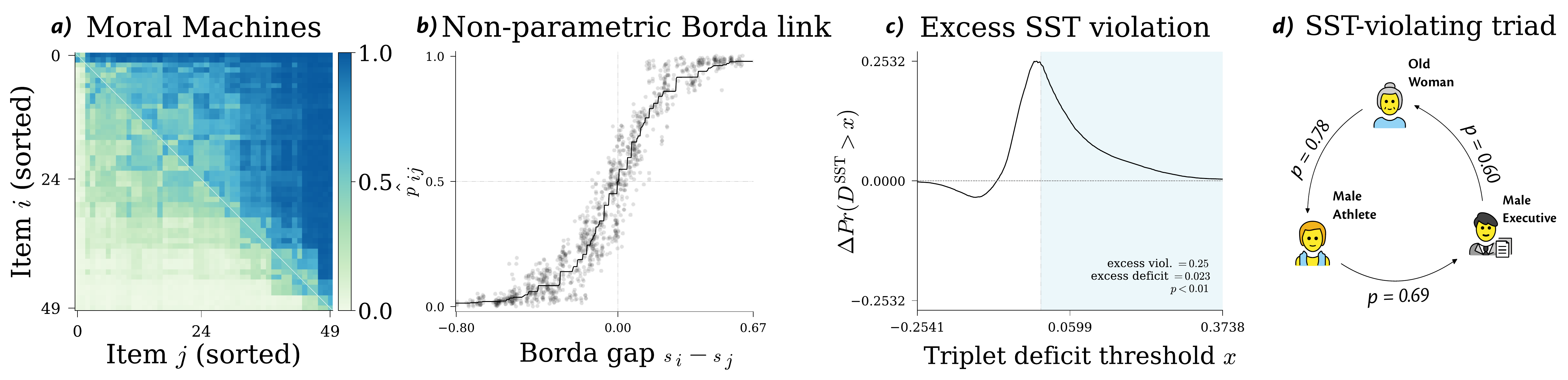}
    \caption{
        Strong stochastic transitivity (SST) violations in \texttt{gpt-oss-20b}'s preferences over the \MoralMachines dataset. 
        (a) Empirical pairwise preference matrix sorted by Borda score.
        Blocks of items whose ranking is not explained by the sorted order are visible, e.g., in the bottom-right corner. 
        (b) A Borda-link null maps score gaps $s_i-s_j$ to pairwise win probabilities without assuming a parametric form.
        (c) Excess upper-tail probability of SST deficits, $\Pr_{\rm obs}(D^{\rm SST}\geq x)-\Pr_{0}(D^{\rm SST}\geq x)$. Positive mass in the violation region indicates that large SST violations are more frequent than under the Borda-link null.
        (d) SST-violating triad generated by \texttt{gpt-oss-20b}. This local example illustrates the type of structured inconsistency measured in panel (c).
    }
    \label{fig:sst-diagnostic}
\end{figure}

\textbf{Sampling noise cannot explain the intransitivity.}
Figure~\ref{fig:sst-diagnostic} illustrates the test for
\GptOss\ on \MoralMachines. The observed upper
tail of SST violations is significantly larger than the null. In the
example shown, the model selects the male athlete for sacrifice over the male
executive, and the male executive over the old woman, but strongly
selects the old woman over the male athlete: an inconsistency no single
ordering can produce. We apply the test to all dataset pairs and reject the null hypothesis in nearly all cases at high significance. Full $D^{\text{SST}}$ values and $p$-values
across all model-dataset pairs are in
Table~\ref{tab:transitivity-controls},
Appendix~\ref{app:transitivity-controls}.

\paragraph{Prompt bias cannot explain intransitivity.}

Several other sources of apparent intransitivity have been proposed, including aggregation over multiple system prompts~\cite{Mizrahi2024,Sclar2024}, order effects~\cite{pezeshkpour2024large,zheng_judging_2023}, and item labels~\cite{Zheng2023a}. In order to rule these out, we recompute the statistics within fixed prompt and presentation-order conditions. We find that even after these controls, the LLM preferences are still significantly intransitive ($p < 0.01$) across all model-data pairs. We report full details with effect sizes in Appendix \ref{app:transitivity-controls}.

The tests show that 1) there are significant levels of SST violation across all models and datasets and 2) in no case can those SST violations be attributed to \emph{any} single set of ordinal preferences under \emph{any} i.i.d. noise model.

\section{Parametric preference model}

The question naturally arises: if intransitivity cannot be attributed to noise, where does it come from?

The literature on social choice theory provides three explanations for how structurally transitive preferences can produce apparent intransitivity:
multi-dimensional comparison~\cite{Tversky1969,Causeur2005,Usami2010},
correlated noise~\cite{Thurstone1927,Tsai2006,Tsai2008,cattelan2012models},
and aggregation effects~\cite{hagele_llulls_2001,Condorcet1785,may_intransitivity_1954}.

We focus on the third: \emph{Aggregation-induced preference intransitivity} occurs when individually transitive preferences combine to produce cycles. The
observation dates back to Ramon Llull's thirteenth-century work on
ecclesiastical elections~\cite{hagele_llulls_2001}. Llull argued
that when sets of transitive preferences are aggregated through
voting, the result can violate transitivity. Condorcet later
formalized this as the eponymous \emph{voting paradox}~\cite{Condorcet1785},
and May extended it to stochastic preferences~\cite{may_intransitivity_1954}.

We propose that LLM preference intransitivity observed in
previous work may be better explained by a mixture of
internally transitive preference orderings. The mechanism is
parallel to the voting case. LLMs are pre-trained on text
produced by millions of authors holding incompatible values,
and post-training aggregates feedback from many annotators
who may apply different value systems to the same comparison.
Both stages embed a population of perspectives in the model
rather than a single coherent agent. We refer to this view as
the \emph{parliament of thoughts}: an LLM treated not as a
single agent with stable preferences, but as a collection of
latent perspectives whose weighted aggregate produces each
answer. If this view is correct, LLM intransitivity should be
better predicted by a model with multiple consistent latent
orderings, each representing a distinct system of value, than
by a single noisy ordering.

Here, we start from the BT model, which defines the win probability as a transformation (via a link function $f$) of the distance in a latent score space.
For the classical BT model, that link function is simply the sigmoid, with $p_{ij}(\bm{\theta}) = f(\theta_i, \theta_j) = \sigma(\theta_i - \theta_j) = e^{\theta_i}/(e^{\theta_i}+e^{\theta_j}).$

We make two extensions to account for the behavior of LLMs.

\textbf{Modeling noisy choices.}
First, LLMs frequently refuse to choose between two alternatives, especially in difficult ethical situations, and, in such cases, may state that they are picking randomly between the two. 
To model this random-choice behavior, we follow the formulation of Jerdee and Newman~\cite{jerdee2024luck}:
With probability $\alpha$, a fair random comparison is made; otherwise, we follow the Bradley--Terry rules.
The noise-augmented win probability is thus
\begin{equation}
    p_{ij}(\alpha, \boldsymbol{\theta}) 
    = \frac{\alpha}{2} + (1-\alpha)\,\sigma(\theta_i - \theta_j).
\end{equation}

\textbf{Mixture extension.} Aggregation-induced intransitivity can be expressed with $K_{\max}$ vectors of latent scores $\{\bm{\theta}_k\}_{k=1}^{K_{\max}}$. A pairwise comparison between items $(i,j)$ is realized by first selecting component $k$ with probability $\pi_k$ and then selecting $i$ with probability $p_{ij}^{(k)} = f(\theta_{ik}, \theta_{jk})$~\cite{pearce2025modeling}.

\textbf{Noise-augmented mixture Bradley--Terry model (MBT).}
Combining these extensions yields the following win probability for component $k$ 
\begin{equation}
    p_{ij}^{(k)}(\alpha_k, \boldsymbol{\theta}_k) = 
        \frac{\alpha_k}{2} + (1-\alpha_k)\,\sigma(\theta_{ik} - \theta_{jk}).
\end{equation}
The complete formulation and fitting procedure follow from the multinomial logit mixture formulation~\cite{pearce2025modeling, train2009discrete}.

To facilitate model fit, we place a Dirichlet process prior with concentration parameter $\gamma$ on the mixture weights and implement it via a truncated stick-breaking representation. 
Per-component scores receive a sum-to-zero Gaussian prior for identifiability, and luck parameters receive a $\mathrm{Beta}(1.5,1.5)$ prior. 
The DP prior has the added benefit of identifying a parsimonious set of mixture components, without requiring additional model selection steps.
Full model specification and inference details appear in Appendix~\ref{app:bt_model}.

\textbf{Validation.} This model is complex and liable to be unidentifiable. 
Hence, we validate the fitting procedure on synthetic data generated from the model's own generative process. 
We assess recovery across varying numbers of ground-truth components, items, comparisons per item, and noise levels, and find that the model recovers parameters well across regimes.
In particular, recovery is faithful for up to four-component models across noise levels and levels of inter-component score correlations, but degrades for five-component models. 
Full recovery results are reported in Appendix~\ref{app:simulation}.

\section{Inferred MBT Components}
\label{sec:mixture_results}
We apply the \MBT\ model to 29 (\texttt{LLM}, \texttt{dataset})
pairs, as illustrated in Figure~\ref{fig:main}, and investigate the number of mixture components needed to
model the preference structure of LLMs in the wild. For posterior draw $s$, we define the effective component count as the perplexity of its mixture weights,
\begin{equation}
    k_{\mathrm{eff}}^{(s)}
    = \exp\!\left[-\sum_{k=1}^{K_{\max}}\pi_k^{(s)}\log \pi_k^{(s)}\right].
\end{equation}
This quantity equals one when one component has all the weight and increases as weight spreads across components. Table~\ref{tab:k_eff} reports the posterior mean of $k_{\mathrm{eff}}^{(s)}$ and its 2.5th and 97.5th percentiles.

  \begin{table}[ht]
      \centering
      \small
      \begin{threeparttable}
      \caption{Effective component count by model and dataset. Entries are the posterior mean of per-draw mixture-weight perplexity, with 95\% credible intervals in brackets.}
      \label{tab:k_eff}
      \begin{tabular}{lcccc}
      \toprule
      Model & \MoralMachines & \HalfLives & \Jokes & \Population \\
      \midrule
      \DeepSeek & 4.89 [4.86, 4.91] & 4.08 [4.06, 4.10] & 1.64 [1.61, 1.67]\tnote{$\dagger$} & 2.99 [2.98, 3.00] \\
      \GptOss & 3.86 [3.81, 3.91] & 1.99 [1.99, 1.99] & 2.00 [1.99, 2.00] & 2.18 [2.17, 2.19] \\
      \GemmaThree & 1.93 [1.93, 1.94] & 1.00 [1.00, 1.00] & 1.00 [1.00, 1.00] & 1.00 [1.00, 1.00] \\
      \GemmaFour & 1.00 [1.00, 1.00] & 1.00 [1.00, 1.00] & 3.12 [3.09, 3.15] & 1.00 [1.00, 1.00] \\
      Human & 2.57 [2.48, 2.67] & --- & --- & --- \\
      \OLMo & 2.43 [2.40, 2.45] & 1.00 [1.00, 1.00] & 3.49 [3.33, 3.83]\tnote{$\dagger$} & 4.86 [4.84, 4.87] \\
      \QwenSmall & 2.81 [2.80, 2.82]\tnote{$\dagger$} & 4.48 [4.47, 4.49] & 1.88 [1.87, 1.89] & 3.03 [3.01, 3.05]\\
      \QwenLarge & 2.85 [2.84, 2.86] & 2.97 [2.94, 3.01]\tnote{$\dagger$} & 1.00 [1.00, 1.01] & 1.11 [1.10, 1.11] \\
      \bottomrule
      \end{tabular}
      \begin{tablenotes}
      \footnotesize
      \item[$\dagger$] Poor HMC convergence: divergent transitions, $\hat R > 1.01$, or ESS $<100$.
      \end{tablenotes}
      \end{threeparttable}
  \end{table}

\subsection{Exploring Inferred Components}
\subsubsection{Value-laden judgments: \MoralMachines\ and \Jokes}

\textbf{\Jokes.} Five of seven models are multimodal on \Jokes,
with $\keff$ ranging from $1.64$ (\DeepSeek) to $3.49$
(\OLMo). The clean prediction that alignment flattens aesthetic
disagreement does not survive: \GemmaFour, the most unimodal
model on \MoralMachines, produces the second-highest \Jokes
$\keff$ ($3.12$). Whatever determines mode structure
in these models is not well-summarized by a moral-versus-
aesthetic axis.

\textbf{\MoralMachines.} 

\begin{figure}[H]
    \centering
    \includegraphics[width=\linewidth]{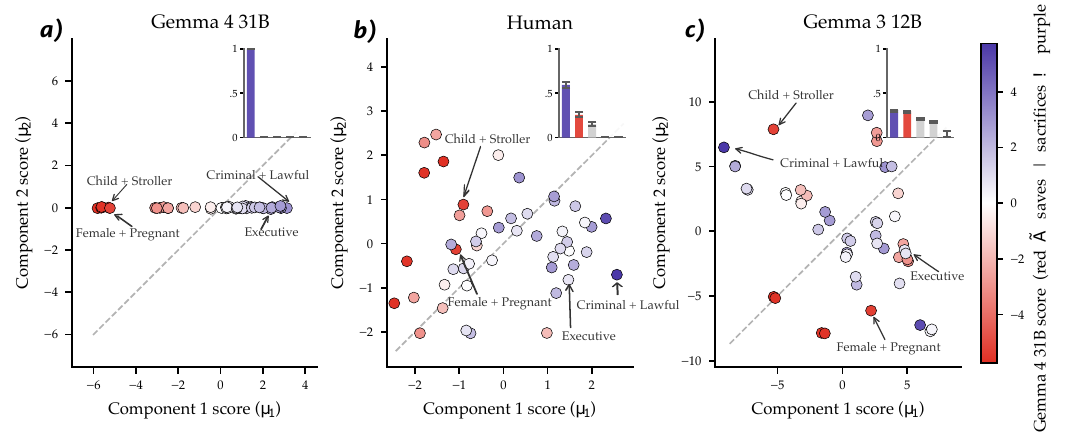}
    \caption{
        \textbf{Two-component posterior scores for \MoralMachines raters.}
Each panel plots Bradley--Terry scores from the two dominant components:
Component~1 on the $x$-axis, Component~2 on the $y$-axis. Items are
colored by their \GemmaFour\ score. Distance from the diagonal measures
disagreement between components: items above the diagonal score higher
in Component~2, those below score higher in Component~1. Bar insets
show posterior mixture weights. For \MoralMachines, the selected person is
the one sacrificed and is encoded as the pairwise winner. Lower scores
therefore indicate greater protection from sacrifice.
    }
    \label{fig:fig_2_posterior}
\end{figure}

We examine \MoralMachines\ in more detail. Figure~\ref{fig:fig_2_posterior} compares the top two mixture components for three raters: (a)~\GemmaFour, (b)~Human, and (c)~\GemmaThree.

The three panels show strikingly different structures. \GemmaFour\ (a) collapses onto a single component (100\%), saving children and pregnant women while sacrificing executives and criminals; we call this the \emph{vulnerability-protective} ethic. Humans (center)
split into three components. The dominant one (59\%) matches
\GemmaFour's vulnerability-protective ethic (Spearman $\rho = 0.68$).
The second (26\%) inverts the preference, saving lawful, high-status
individuals (doctors, executives) at the expense of children and
pregnant women; we call this the \emph{status-protective} ethic. A
third minor component (15\%, not plotted) bears no relation to
\GemmaFour\ ($\rho \approx 0$). \GemmaThree\ (right) is the most
internally heterogeneous, with the widest off-diagonal spread and
largest axis range. Its components resist clean interpretation as
either vulnerability- or status-protective: Component~2 saves
Female+Pregnant yet sacrifices Child+Stroller, and Executive sits
midway on both axes.

Three findings stand out. First, the mixture exposes a clear
multimodality in human preferences that is hidden in the aggregate:
while the dominant vulnerability-protective ethic accounts for 59\%
of comparisons, a sizable minority of 26\% come from a
status-protective component. Second, \GemmaFour\ aligns solely with
the dominant vulnerability-protective human mode, while
\GemmaThree's components match neither the vulnerability- nor the
status-protective ethic. Third, despite both being from the same model family, the two Gemma models diverge sharply: aggregate posterior
predictive rankings and Borda counts give a Spearman correlation of
$\rho = 0.06$ between them. The newer and larger \GemmaFour\
succeeds at alignment where the earlier, smaller model did not,
consistent with prior findings that preference alignment scales
with model size~\cite{mazeika_utility_2025,Takemoto2026}. Taken as a whole,
these results argue that LLM preference patterns can reflect the
diversity of human morality while still escaping simple
categorization through human preference modes.
\subsubsection{Factual judgments: \Population\ and \HalfLives}

\begin{wrapfigure}{r}{0.5\linewidth}
    \hspace{3em}
    \centering
    \includegraphics[width=0.49\textwidth]{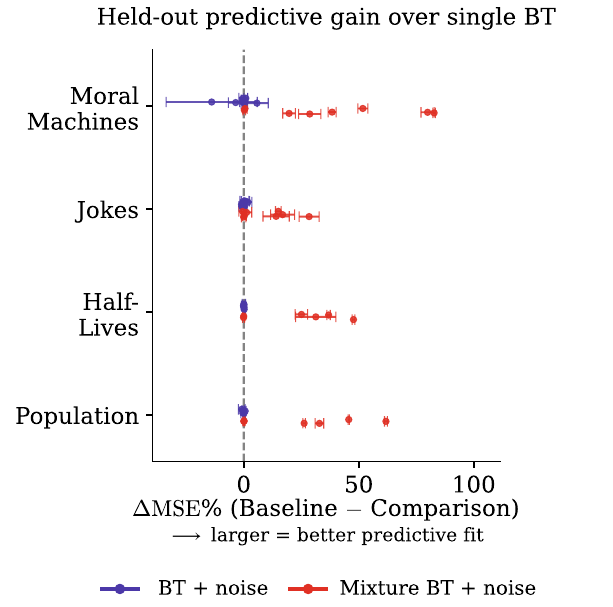}
    \caption{\textbf{Held-out predictive accuracy measured by mean squared error (MSE).} We compare a noise-augmented single-component Bradley--Terry model and the \MBT\ model against the standard Bradley--Terry baseline. Points show $\DeltaMSE$ across model--dataset pairs; horizontal intervals show variation across the five folds. Values to the right of zero indicate better held-out prediction than the baseline. The \MBT\ model generally matches or improves on single-component baselines, with the largest gains on \MoralMachines\ and \HalfLives, where posterior fits also indicate multiple effective components.
}
    \vspace{-2em}
    \label{fig:cv}
\end{wrapfigure}

While factual tasks like estimating city population or half-lives should produce no intransitivity in a well-calibrated model, we observe it across
most models we test. While \GemmaThree, \GemmaFour, and \QwenLarge\
recover a single component on population, the remaining four
models all show multimodality, with effective component counts
between $2.18$ (\GptOss) and $4.86$ (\OLMo).

The mixture structure reveals that this reflects structured confusion about specific items rather than uniform noise.
\DeepSeek\ ranks Denver (the 19th largest US city) as \#3 in
components 1 and 2 (mixture weights 29\% and 12\%), nearly tied
with New York, but \#41 in the dominant component; 13.8\% of
triplets involving Denver violate strong stochastic transitivity,
against a dataset average of 5.7\%. \QwenSmall\ shows the same
pattern for Louisville (24th largest), and \DeepSeek\ inflates
Detroit inconsistently across components.
Figure~\ref{fig:population_intransitivity} in
Appendix~\ref{app:population_data} visualizes these patterns.

\HalfLives\ shows the same shape, more sharply: four models
produce multimodal preferences (up to $\keff = 4.48$
for \QwenSmall), while three (\GemmaThree, \GemmaFour, \OLMo)
recover a single component. This is consistent with a difficult
factual task where models with partial knowledge get persistently
confused across several modes, while those without it collapse
onto a single uninformative ranking.

Multimodality therefore arises from two distinct sources. In value-laden tasks it reflects value pluralism, with components representing coherent but distinct systems of value, consistent
with a social-choice reading of LLMs as a parliament of thoughts
whose decisions resemble voting more than individual
deliberation. In factual tasks it reflects structured confusion:
persistent, item-specific errors rather than uniform noise.

  \section{MBT outpredicts single-utility Bradley--Terry}

  The \MBT\ model with $K_{\max}=5$ components has more parameters than a single Bradley--Terry baseline, raising the question of whether gains reflect overfitting rather than genuine heterogeneity. We rule this out with 5-fold cross-validation across all four tasks and all LLMs. We randomly assign each item pair to one of five folds and hold out all comparisons for that pair at test time, fitting on the remaining four. We compare three variants: $K_{\max}=1$ without noise augmentation (the standard Bradley--Terry baseline), $K_{\max}=1$ with noise augmentation, and \MBT\ with $K_{\max}=5$. For either comparison model, we define
  \begin{equation}
      \DeltaMSE
      = \mathrm{MSE}_{\mathrm{BT}} - \mathrm{MSE}_{\mathrm{model}}.
  \end{equation}
  Positive values favor the comparison model over standard Bradley--Terry.

  Figure~\ref{fig:cv} shows $\DeltaMSE$ across all LLM--task pairs. The \MBT\ model matches or outperforms both baselines in the majority of cases. Gains are largest on \HalfLives and \MoralMachines, where $\keff \gg 1$; on \Population and \Jokes, where posteriors are near-unimodal, the advantage is small or absent. The \MBT\ model is favored exactly where the data exhibit the structure it is designed to capture.

\section{Aggregation flattens underlying preference heterogeneity}
\label{sec:hetero}

Judge ensembles, simple reward models, and arena leaderboards aggregate heterogeneous model behavior to a single ranking. While these are often instructive, they mask underlying heterogeneity; models which display distinct preference modes can appear similar in aggregate. Understanding the mechanisms driving preferences therefore requires examining the disaggregated modes. The aggregate is an average over coherent internal modes, and two such averages can match or diverge for reasons unrelated to whether the underlying preferences match.

In Figure~\ref{fig:component_het} we show that aggregation obscures both similarity and difference between preference modes. We show the aggregate Spearman correlation (black bars) with the individual component-component correlation (circles, sized and colored by combined weight $\pi_i \pi_j$).
These results illustrate a few salient categories:

\begin{wrapfigure}{r}{0.5\linewidth}
    \vspace{-0.8em}
    \includegraphics[width=0.47\textwidth]{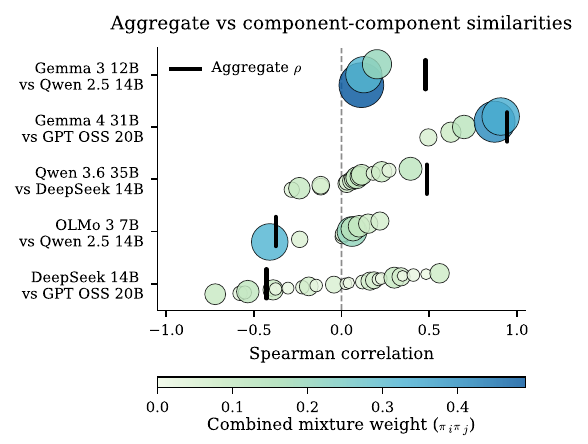}
    \caption{\textbf{Aggregation hides heterogeneous cross-model structure.} For selected \MoralMachines\ model pairs, horizontal black bars show the Spearman correlation between the two aggregate preference matrices. Circles show Spearman correlations between individual \MBT\ components across the same two models. Circle size and color encode the combined mixture weight $\pi_i \pi_j$, so larger, darker points correspond to component matches that contribute more to the aggregate comparison. Aggregate similarity can obscure both aligned and misaligned latent components.}
    \label{fig:component_het}
    \vspace{-3em}
\end{wrapfigure}

\textbf{1) Aggregate agreement with component disagreement.} Two models can have similar aggregate rankings while some component pairs disagree (\QwenLarge $\times$ \GptOss) or have little similarity (\GemmaThree $\times$ \QwenSmall). \textbf{2) Aggregate disagreement with component overlap.} Two models can have dissimilar aggregate rankings while some component pairs agree (\DeepSeek $\times$ \GptOss). \textbf{3) Agreement at both levels.} Similar aggregate rankings can coincide with similar component rankings (\GemmaFour $\times$ \GptOss). \textbf{4) Disagreement at both levels.} Dissimilar aggregate rankings can coincide with broadly dissimilar component rankings (\OLMo $\times$ \QwenSmall). Together, these patterns show what aggregate model comparisons can omit.

\section{Limitations}
Three limitations remain. Correlated noise across item comparisons is not separately identified from multimodality without explicit covariance modeling~\citep{cattelan2012models}. Our parametric link could be replaced with an MLP or Gaussian process to absorb covariates like prompt order. The inferred components lack independent validation; topic modeling on reasoning traces or mechanistic interpretability of
activations would directly strengthen the structural claim. We leave these to future work.

\section{Conclusion}
LLM revealed preferences are often evaluated as if they come from a single latent ordering. 
Our results show this assumption is incorrect. 
Across models and domains, observed pairwise preferences are intransitive in ways no single global ordering can explain. 
This does not entail that the underlying preferences are structurally intransitive: A mixture of internally coherent orderings can produce the same surface pattern. 
Our \MBT\ model formalizes this alternative, and cross-validation shows that multi-component fits often improve held-out prediction. LLM preferences are better understood as containing multiple latent orderings rather than as noisy realizations of a single one. 
Finally, we unmask LLM preferences to show how the aggregates flatten underlying preference heterogeneity.

This paper draws inspiration from social choice theory, a body of work developed to understand collective governance through voting and elections. We conceptualize LLMs not as monolithic, coherent agents but as a parliament of thoughts. We have shown that aggregate LLM preferences hide a latent multiplicity, and we suggest that alignment may be a question of governance. Our framework provides both qualitative evidence for this view and a quantitative toolkit for inferring, understanding, and ideally controlling preference modes, enabling targeted interventions toward safer AI.

\bibliographystyle{plainnat}
\bibliography{refs}

\newpage
\appendix

\section{Dataset preparation}
\label{app:data-prep}

This appendix describes how we arrive at each processed set of items discussed in Section~\ref{sec:pairwise_pref_llm}.
Across these datasets, we preserve exactly $50$ items so that the four problems present LLMs with the same number of alternatives and unordered pairs.
This maximum is set by the \MoralMachines\ dataset, where human data are limited to $60$ items.
For datasets in which the preference structure is derived from human pairwise comparisons, we restrict our initial subset to the giant strongly connected component of the human preference graph.
This guarantees that Bradley--Terry rankings are identifiable even in the absence of strong regularization~\cite{newman2023efficient}.
The SCC condition is automatically met for datasets in which comparisons follow from an external, factual score (\Population, \HalfLives).

\paragraph{\Jokes.}
The initial dataset comprises $4,030$ jokes with crowd-sourced ratings.
The ratings are elicited with pairwise comparisons, arranged so that each joke appears in exactly 14 unique pairs and each pair is rated by 5 annotators~\cite{simpson_predicting_2019}.
Ratings can be one of \texttt{A}, \texttt{B}, or \texttt{X} (``neither funnier''), and we keep only the definite ratings when constructing the preference matrix.
This removes a fraction of the comparisons but does not remove any jokes from the dataset, since each joke happens to retain at least one decisive rating.
The resulting directed preference graph is a single strongly connected component; we select a random subset of $50$ jokes, and use the same subset throughout our analysis.

\paragraph{\MoralMachines.}
The Moral Machine dataset~\cite{awad2018moral} captures the decisions humans think a self-driving car ought to make in a wide variety of scenarios involving an imminent collision.
We restrict our analysis to one set of scenarios, in which the victims' attributes are randomized, the number of characters is equal, and both alternatives are set to one.
This selection isolates the categorical attributes of pedestrians (age, profession, gender, social status, legality of crossing) from quantity-based assessments.
A crossing signal is included in some scenarios and determines whether cars and pedestrians are obeying the law.
Each unique \texttt{(pedestrian category, crossing signal)} combination defines an item, yielding 60 distinct items.
Aggregating human votes into a directed preference graph and computing strongly connected components leaves two strongly connected components of $54$ and $6$ items, respectively; we work with the larger one and take a fixed random subset of $50$ as our experimental set.

\paragraph{\Population.}
We use the U.S.~Census Bureau estimates (table \texttt{SUB-IP-EST2023-POP}), which lists $19,484$ U.S.~cities with an April 2020 baseline population and July 1\textsuperscript{st} estimates for each year from 2020 to 2023.
We score each city by its 2023 estimate, the most recent value and the one most likely to be reflected in LLM training data.
Naively taking the top 50 cities yields several adjacent pairs whose populations differ by only a few hundred residents; these differences are within Census estimation uncertainty, and we do not want to test LLMs on them.
Hence, we apply a minimum gap filter and sort cities by population in descending order, greedily skipping any city whose population is within $3,000$ of the previously selected city, and we stop after $50$ selections.
Deterministic pairwise comparisons are then written to a dense matrix comparing these cities based on the 2023 estimate.

\paragraph{\HalfLives.}
All known nuclides are downloaded from the IAEA LiveChart of Nuclides API~\cite{iaea_livechart} (using the \texttt{ground\_states} endpoint).
We discard entries with non-numeric or missing half-lives, including isotopes flagged as stable, and keep only ground-state nuclides with parsable atomic number, mass number, and half-life in seconds.
This yields the candidate pool of all known radioactive isotopes with measured half-lives.
From this pool we draw a random sample of $100$ isotopes, then sort them by descending half-life to define a deterministic item order.
Our final item set comprises the first 50, i.e., the 50 longest-lived isotopes in the random sample.
Deterministic pairwise comparisons are then written to a dense matrix comparing the isotopes based on their half-lives.

\section{Language models}
\label{app:llm-provenance}

This appendix documents the language models used in our pairwise preference elicitation experiments. 
All models are open-weight models accessed through public model repositories. 
We use open-weight models to make the elicitation protocol reproducible and to avoid dependence on hosted APIs whose model weights, system behavior, or decoding defaults may change over time.

\subsection{Model selection}

We selected models to cover several model families, sizes, and post-training styles while keeping inference feasible for dense pairwise elicitation. 
The final set contains seven instruction-tuned or reasoning-oriented models: \DeepSeek, \GptOss, \GemmaFour, \GemmaThree, \OLMo, \QwenLarge, and \QwenSmall. 
For each model, we use the released checkpoint without additional fine-tuning.

\subsection{Model provenance}

Table~\ref{tab:llm-provenance} lists the exact model identifiers used in our experiments. 
When a peer-reviewed paper or technical report is available, we cite it as the primary source. 
When the exact checkpoint is not covered by a formal paper, we cite the public model card or release page.
\begin{table}[t]
\centering
\small
\setlength{\tabcolsep}{4pt}
\caption{
Language-model provenance for pairwise preference elicitation.
Model identifiers correspond to the exact checkpoints used in our experiments.
Citations point to the model technical report or model card used to document provenance.
}
\label{tab:llm-provenance}
\begin{tabularx}{\linewidth}{lXl}
\toprule
Display name & Model identifier & License / terms \\
\midrule
\DeepSeek 
& \texttt{deepseek-ai/\DeepSeek-R1-Distill-Qwen-14B}~\citep{deepseekai2025deepseekr1incentivizingreasoningcapability}
& MIT \\

\GptOss 
& \texttt{openai/gpt-oss-20b}~\citep{openai2025gptoss120bgptoss20bmodel}
& Apache 2.0 \\

\GemmaThree 
& \texttt{google/gemma-3-12b-it}~\citep{gemma_2025}
& Gemma terms \\

\GemmaFour 
& \texttt{google/gemma-4-31b-it}~\citep{google_deepmind_gemma4_model_card_2026}
& Gemma terms \\

\OLMo Think 
& \texttt{allenai/Olmo-3-7B-Think}~\citep{olmo2025olmo3}
& Apache 2.0 \\

\QwenSmall 
& \texttt{Qwen/Qwen2.5-14B-Instruct}~\citep{qwen2.5, qwen2}
& Apache 2.0 \\

\QwenLarge 
& \texttt{Qwen/Qwen3.6-35B-A3B}~\citep{qwen36_35b_a3b}
& Apache 2.0 \\
\bottomrule
\end{tabularx}
\end{table}
\subsection{Inference setup}

For every model--dataset pair, we use VLLM to query the model on all unordered item pairs under the prompt and order controls described in Appendix~\ref{app:preference-protocol}. 
We use the released instruction or reasoning checkpoint as-is and do not perform any additional training, calibration, or preference tuning. 
Each generation is parsed only for the final forced-choice identifier. 
For reasoning models, any reasoning trace is not used as data except insofar as it precedes the final parsed choice.

Unless otherwise stated, decoding follows the model configuration used for the elicitation run. 
The same elicitation protocol is applied across models: identical item sets, identical prompt variants, randomized pseudo-identifiers, both presentation orders, and the same parsing rules. 
Unparseable outputs are discarded before constructing the pairwise win-count matrix.

\section{Preference elicitation protocol}
\label{app:preference-protocol}

This appendix describes the full elicitation and parsing procedure used to construct the pairwise preference matrix from LLM responses. 
At a high level, for every unordered pair of items in a dataset, we generate dense comparisons under randomized presentation orders, with prompt variation and randomized pseudo-identifiers.

\textbf{System prompt variants} are designed to preserve the same task semantics while changing surface wording. 
For example, in the jokes domain, the model is always asked to choose the funnier joke, while in the Moral Machine domain it is always asked to choose the person who should be sacrificed.
We generate four variations for each dataset by including or excluding a roleplay prompt and varying the wording of the comparison.
For instance, when experimenting on \textsc{jokes}, the system prompt begins with one of the following options:
\begin{itemize}
    \item (\texttt{roleplay, comparison semantic 1}). You are a professional comedian with refined taste in humor. You will be given two jokes labeled \texttt{pseudo\_id\_A} and \texttt{pseudo\_id\_B}. Select the funnier one.
    \item (\texttt{roleplay, comparison semantic 2}). You are a comedian with a lot of professional experience evaluating material for a stand-up show. You will be given two jokes labeled \texttt{pseudo\_id\_A} and \texttt{pseudo\_id\_B}. Determine which joke is funnier.
    \item (\texttt{no roleplay, comparison semantic 1}). You will be given two jokes labeled \texttt{pseudo\_id\_A} and \texttt{pseudo\_id\_B}. Select the funnier one.
    \item (\texttt{no roleplay, comparison semantic 2}). You will be given two jokes labeled \texttt{pseudo\_id\_A} and \texttt{pseudo\_id\_B}. Determine which joke is funnier.
\end{itemize}
It is followed by boilerplate prompting that encourages structured output.

\textbf{Ordering and labeling bias} is mitigated by presenting items in both orderings an equal number of times, and replacing the pseudo-identifiers with randomly sampled label letters at inference time. 
We use a simple user message formatted as
\begin{equation*}
    \texttt{<pseudo\_id\_A>: <item A>} \qquad
    \texttt{<pseudo\_id\_B>: <item B>}.
\end{equation*}

\textbf{Sampling}.
Comparisons are sampled $M$ times for each comparison at the default temperature specified in the model configuration.
For thinking models, the answer is parsed after the reasoning trace; for non-thinking models, the answer is parsed directly from the generation. 
An output is included only if the selected value matches one of the two pseudo-identifiers assigned to the current comparison. 
Otherwise, it is discarded as unparsable.
This produces a successful match count that can vary slightly across item pairs if some generations fail to parse.

\begin{algorithm}[H]
\caption{Pairwise preference elicitation}
\label{alg:preference-elicitation}
\begin{algorithmic}[1]
\Require Comparison items $[N]=\{1,\ldots,N\}$, prompt variants $\mathcal{V}$, samples per comparison $M$
\State Initialize win-count matrix $A\leftarrow 0_{N\times N}$
\For{$i=1$ to $N$}
    \For{$j=i+1$ to $N$}
        \State Sample two distinct pseudo-identifiers $u,v$ without replacement from $\{A,B,...,Z\}$
        \For{each prompt variant $s\in\mathcal{V}$}
            \For{each order $o\in\{(i,j),(j,i)\}$}
                \State Construct a forced-choice prompt with items in order $o$
                \State Query the model $M$ times
                \For{each generated completion}
                    \State Extract the value of the selected item, \texttt{selected}
                    \If{\texttt{selected} matches the pseudo-identifier assigned to item $i$}
                        \State $a_{ij}\leftarrow a_{ij}+1$
                    \ElsIf{\texttt{selected} matches the pseudo-identifier assigned to item $j$}
                        \State $a_{ji}\leftarrow a_{ji}+1$
                    \Else
                        \State Discard the generation as unparsable
                    \EndIf
                \EndFor
            \EndFor
        \EndFor
    \EndFor
\EndFor
\State \Return win-count matrix $A$
\end{algorithmic}
\end{algorithm}

\section{Transitivity Violation Null Model}
\label{app:null-model}
\subsection{Notation}

For each pair of items $(i, j)$, let $a_{ij}$ denote the number of
times $i$ was selected over $j$ in the LLM's pairwise comparisons,
and let $\hat{p}_{ij} = a_{ij} / (a_{ij} + a_{ji})$ be the empirical
win probability. The \emph{Borda count} of item $i$ is the average
win probability against all other items in the dataset,
\begin{equation}
b_i = \frac{1}{N-1}\sum_{j \neq i} \hat{p}_{ij},
\end{equation}
where $N$ is the number of items.

\subsection{Borda-link null}
\label{app:borda-null}

A random utility model assumes $p_{ij} = f(\theta_i - \theta_j)$ for
some non-decreasing link $f$ and latent scores
$\{\theta_i\}_{i=1}^N$. Our null tests whether the observed
comparisons could have come from any such model, without committing
to a particular $f$ or inferring the $\theta_i$.

\paragraph{Borda count as a monotone proxy.}
Treating the latent scores $\theta_j$ for $j \neq i$ as draws from a
 distribution $p$ with mean $\bar\theta$ and variance
$\sigma^2$, the conditional expected Borda count is the convolution
of the link with the score distribution,
\begin{equation}
\mathbb{E}\!\left[b_i \mid \theta_i\right]
= \mathbb{E}_{\theta \sim p}\!\left[f(\theta_i - \theta)\right]
= (f * p)(\theta_i).
\end{equation}
Writing $V = \theta_i - \theta$, so $\mathbb{E}[V] = \theta_i - \bar\theta$
and $\mathrm{Var}(V) = \sigma^2$, and Taylor-expanding $f$ around the
centered score $\theta_i - \bar\theta$,
\begin{equation}
\mathbb{E}\!\left[b_i \mid \theta_i\right]
= f(\theta_i - \bar\theta)
  + \tfrac{\sigma^2}{2}\, f''(\theta_i - \bar\theta)
  + O(\sigma^3).
\end{equation}
For score distributions concentrated around their mean ($\sigma$
small relative to the curvature scale of $f$), the leading term
dominates and the map
$\theta_i \mapsto \mathbb{E}[b_i \mid \theta_i]$ is monotone in
$\theta_i$. Any monotone function of the score gap
$\theta_i - \theta_j$ can therefore be expressed as a monotone
function of the Borda gap $b_i - b_j$, which is the only structure
the isotonic null below requires
\paragraph{Isotonic null.}
Motivated by this approximation, we take the null win probability to
be a non-decreasing function of the Borda gap,
\begin{equation}
p^{\text{null}}_{ij} = g(b_i - b_j),
\end{equation}
and fit $g$ by isotonic regression on the observed pairs
$(b_i - b_j,\; \hat{p}_{ij})$, anchored at $g(0) = 1/2$ to enforce
symmetry. This yields the closest random-utility-consistent fit to
the observed comparisons without committing to a parametric link.

\subsection{Bootstrap test}
\label{app:bootstrap-test}

Under the null, the wins follow
$a^{\text{null}}_{ij} \sim \mathrm{Binomial}(M_{ij},\,
p^{\text{null}}_{ij})$, where $M_{ij} = a_{ij} + a_{ji}$ is the total
number of comparisons between $i$ and $j$. We generate
$B = 1000$ bootstrap samples $\{a^{(b)}_{ij}\}_{b=1}^B$, recompute
the SST deficit $D^{\text{SST}}_{(b)}$ on each, and report the
upper-tail probability
\begin{equation}
\hat{P}_{\text{null}}\!\big(x \geq D^{\text{SST}}_{\text{obs}}\big)
= \frac{1}{B}\sum_{b=1}^B
  \mathbb{1}\big\{D^{\text{SST}}_{(b)} \geq
                    D^{\text{SST}}_{\text{obs}}\big\}.
\end{equation}
A small upper-tail probability indicates that the observed deficit is
larger than what any random utility model would produce.

The aggregate diagnostics in Section~\ref{sec:structured-inconsistency} could be inflated by protocol artifacts. 
For example, a model may prefer the first option, respond differently to different system-prompt phrasings, or interact with pseudo-labels in a non-item-specific way. We therefore repeat the transitivity diagnostics under several restricted conditions.

\section{Transitivity Controls}\label{app:transitivity-controls}

\paragraph{Full aggregate matrix.}
The primary analysis pools all successful completions for each unordered pair $\{i,j\}$. 
This produces the aggregate matrix $A$ used in the main text. 
It has the largest number of observations per pair, but it also pools across prompt variants and presentation orders.

\paragraph{Fixed prompt and fixed order.}
To test whether triplet residuals are caused by pooling across prompt
conditions, we recompute the diagnostics within each fixed prompt condition.
We further split by presentation order. For an unordered pair $\{i,j\}$ with $i<j$, one condition contains prompts where the lower-index item appears in slot A, and the other contains prompts where it appears in slot B. This gives condition-specific matrices of the form $A^{(c)},$ where $c$ indexes a prompt-order condition.

For each condition, we recompute the Borda scores, refit the monotone
Borda-link null, and regenerate a bootstrap distribution using only the
counts available in that condition. Thus each condition has its own global
ordering, comparison counts, and null calibration.

\paragraph{Order-symmetrized matrix.}
We also construct an order-symmetrized matrix that removes first-order
presentation bias. For each unordered pair $\{i,j\}$ with $i<j$, define
\[
    \hat p^{A}_{ij}
    =
    \Pr(i\succ j \mid i\text{ shown in slot A}),
\]
and
\[
    \hat p^{B}_{ij}
    =
    \Pr(i\succ j \mid i\text{ shown in slot B}).
\]
When both orientations are available, we estimate the order-symmetrized
preference probability as
\[
    \hat p^{\mathrm{sym}}_{ij}
    =
    \frac{1}{2}
    \left(
        \hat p^{A}_{ij}
        +
        \hat p^{B}_{ij}
    \right).
\]
If only one orientation is available, we use the available estimate. We then
construct a symmetrized count matrix with the same total number of successful
comparisons for the pair:
\[
    a^{\mathrm{sym}}_{ij}
    =
    m_{ij}\hat p^{\mathrm{sym}}_{ij},
    \qquad
    a^{\mathrm{sym}}_{ji}
    =
    m_{ij}\left(1-\hat p^{\mathrm{sym}}_{ij}\right).
\]
This matrix is used only for diagnostics, not as raw observed count data.

\paragraph{Pseudo-label controls.}
Pseudo-identifiers are randomly assigned to options and are independent of item identity. In the main experiments, they are used only to parse the selected option. If pseudo-label effects are a concern, the same diagnostic can be repeated within fixed pseudo-label conditions. In the current analysis, we focus on prompt and presentation-order controls because they are the most direct protocol-level explanations for transitivity artifacts.

\paragraph{Control statistic.}
For each control condition $c$, we compute
\[
    \Delta T^{(c)}
    =
    T(\hat P^{(c)})
    -
    \mathbb{E}_{0}
    \left[
        T(\hat P^{*(c)})
    \right],
\]
where the null distribution is fitted and bootstrapped separately within
condition $c$. We report the same WST, SST, and MST excesses as in the main
analysis.

\paragraph{Interpretation.}
If excess triplet residuals disappear under fixed prompt or order-symmetrized conditions, then the aggregate result is likely driven by a protocol artifact. If the residuals persist within fixed conditions, then they are less likely to be explained by prompt wording or presentation order
alone. This does not identify the source of the structure, but it strengthens
the case that the residuals are not merely random sampling noise around one
global ordering.

\begin{table}[H]
    \centering
    \small
    \setlength{\tabcolsep}{4pt}
    \caption{
    Triplet-level transitivity diagnostics under fixed prompt and order controls.
    For each model--dataset pair, diagnostics are recomputed within each fixed system-prompt and presentation-order stratum
    ($2$ prompt styles $\times$ $2$ semantic variants $\times$ $2$ presentation orders).
    Entries report the median excess across the eight strata, with the minimum stratum-level excess in parentheses.
    Excess is defined as the observed statistic minus the bootstrap mean under a matched Borda-link null fitted within the same stratum.
    SST uses mean positive triplet deficit; MST uses mean absolute multiplicative cycle residual.
    Positive minima indicate that excess intransitivity appears in every fixed prompt--order stratum.
    All fixed-stratum tests are significant at $p \leq .05$; most attain the bootstrap floor $p=.001$.
    }
    \label{tab:transitivity-controls}
    \begin{tabular}{llcc}
    \toprule
    Model & Dataset & $\Delta$SST & $\Delta$MST \\
    \midrule
    \multirow{4}{*}{\GptOss}
    & \Jokes & 0.0115 (0.0110) & 0.0165 (0.0156) \\
    & \MoralMachines & 0.0259 (0.0201) & 0.0291 (0.0229) \\
    & \Population & 0.0167 (0.0129) & 0.0075 (0.0060) \\
    & \HalfLives & 0.0192 (0.0177) & 0.0241 (0.0226) \\
    \midrule
    \multirow{4}{*}{\DeepSeek}
    & \Jokes & 0.0240 (0.0187) & 0.0321 (0.0268) \\
    & \MoralMachines & 0.0776 (0.0719) & 0.0726 (0.0578) \\
    & \Population & 0.0168 (0.0133) & 0.0143 (0.0102) \\
    & \HalfLives & 0.0214 (0.0200) & 0.0299 (0.0284) \\
    \midrule
    \multirow{4}{*}{\GemmaFour}
    & \Jokes & 0.0113 (0.0095) & 0.0115 (0.0101) \\
    & \MoralMachines & 0.0329 (0.0235) & 0.0313 (0.0221) \\
    & \Population & 0.0112 (0.0035) & 0.0116 (0.0036) \\
    & \HalfLives & 0.0308 (0.0271) & 0.0308 (0.0258) \\
    \midrule
    \multirow{4}{*}{\GemmaThree}
    & \Jokes & 0.0275 (0.0243) & 0.0282 (0.0236) \\
    & \MoralMachines & 0.0864 (0.0792) & 0.0813 (0.0761) \\
    & \Population & 0.0241 (0.0147) & 0.0243 (0.0146) \\
    & \HalfLives & 0.0414 (0.0273) & 0.0350 (0.0203) \\
    \midrule
    \multirow{4}{*}{\OLMo}
    & \Jokes & 0.0577 (0.0470) & 0.0629 (0.0534) \\
    & \MoralMachines & 0.0924 (0.0739) & 0.0931 (0.0729) \\
    & \Population & 0.0633 (0.0368) & 0.0647 (0.0349) \\
    & \HalfLives & 0.0771 (0.0711) & 0.0777 (0.0711) \\
    \midrule
    \multirow{4}{*}{\QwenSmall}
    & \Jokes & 0.0379 (0.0322) & 0.0361 (0.0308) \\
    & \MoralMachines & 0.0393 (0.0131) & 0.0244 (0.0077) \\
    & \Population & 0.0303 (0.0177) & 0.0300 (0.0170) \\
    & \HalfLives & 0.0551 (0.0510) & 0.0524 (0.0461) \\
    \midrule
    \multirow{3}{*}{\QwenLarge}
    & \Jokes & 0.0172 (0.0156) & 0.0225 (0.0210) \\
    & \MoralMachines & 0.0859 (0.0792) & 0.0807 (0.0774) \\
    & \Population & 0.0002 (0.0002) & 0.0002 (0.0001) \\
    \bottomrule
    \end{tabular}
\end{table}

\section{Noise-augmented mixture Bradley--Terry model}
\label{app:bt_model}

The data are $m$ comparisons among items $[N]=\{1,\ldots,N\}$. For trial $t$, let $(i_t,j_t)$ be the ordered pair shown to the model and let $y_t=1$ if the model selects $i_t$ and $y_t=0$ if it selects $j_t$. Thus
\begin{equation}
    \mathcal D=\{(i_t,j_t,y_t)\}_{t=1}^{m}.
\end{equation}

\subsection{Noise augmentation}
\label{app:luck}

We model LLMs as approaching difficult choices as a toss-up, using a two-step stochastic process~\cite{newman2023efficient}. 
We draw a binary random variable $U_t \sim \mathrm{Bernoulli}(\alpha)$; if $U_t = 1$, the model chooses between the two items with a fair coin flip and, if $U_t = 0$, it follows Bradley--Terry probabilities.
Thus, for comparison $t$,
\begin{align}
    U_t &\sim \mathrm{Bernoulli}(\alpha), \\
    Y_t \mid U_t=1 &\sim \mathrm{Bernoulli}(1/2), \\
    Y_t \mid U_t=0 &\sim \mathrm{Bernoulli}\!\left(\frac{\lambda_{i_t}}{\lambda_{i_t}+\lambda_{j_t}}\right).
\end{align}
where $\theta_i \in \mathbb{R}$ is the latent score of item $i$ and $\lambda_i = e^{\theta_i}$ is its strength.
Marginalizing over $U_t$ gives the win probability
\begin{equation}
    p_{ij}(\alpha, \boldsymbol{\theta}) 
    = \frac{\alpha}{2} + (1-\alpha)\,\frac{\lambda_i}{\lambda_i + \lambda_j}.
\end{equation}
\subsection{MBT model}
\label{app:mixture_w_luck}
We model the multiplicity of ways a model can rank items as a mixture, adding a step to the stochastic process.
This requires $K_{\max}$ vectors of latent scores $\{\bm{\theta}_k\}_{k=1}^{K_{\max}}$ and associated luck parameters $\{\alpha_k\}_{k=1}^{K_{\max}}$.
The mixture component selected for comparison $t$ is then modeled as a categorical variable
\begin{align}
    Z_t &\sim \mathrm{Cat}(\boldsymbol{\pi}), \\
    Y_t \mid Z_t = k &\sim \mathrm{Bernoulli}\bigl(p_{i_tj_t}^{(k)}\bigr).
\end{align}
where
\begin{equation*}
    p_{ij}^{(k)}(\alpha_k, \bm{\theta}_k)
    = \frac{\alpha_k}{2} + (1-\alpha_k)\,\frac{\lambda_{ik}}{\lambda_{ik}+\lambda_{jk}},
\end{equation*}
and $\lambda_{ik} =e^{\theta_{ik}}$ is the strength of item $i$ in component $k$.

\subsection{Complete-data and marginal likelihood}
\label{app:complete}

The joint likelihood of the comparisons $\mathcal{D}$ and assignments $\bm{Z}$, written using $z_{tk}=\mathbb{1}\{Z_t=k\}$, is
\begin{equation}
    \Pr(\mathcal{D}, \boldsymbol{Z} \mid \boldsymbol{\theta}, \boldsymbol{\pi}, 
    \boldsymbol{\alpha}) = 
    \prod_{t=1}^{m} \prod_{k=1}^{K_{\max}}
    \left[\pi_k
    \bigl(p_{i_tj_t}^{(k)}\bigr)^{y_t}
    \bigl(1-p_{i_tj_t}^{(k)}\bigr)^{1-y_t}\right]^{z_{tk}}
\end{equation}
Marginalizing over the component assignments $\bm{Z}$ yields
\begin{equation}
    \Pr(\mathcal D\mid\boldsymbol\theta,\boldsymbol\pi,\boldsymbol\alpha)
    = \prod_{t=1}^{m}\left[\sum_{k=1}^{K_{\max}}\pi_k
    \bigl(p_{i_tj_t}^{(k)}\bigr)^{y_t}
    \bigl(1-p_{i_tj_t}^{(k)}\bigr)^{1-y_t}\right]
\end{equation}
which can be re-expressed in terms of the model's sufficient statistics, namely the number of wins by $i$ over $j$ for all pairs.
Let
\begin{equation}
    a_{ij}=\sum_{t=1}^{m}
    \left[
    \mathbb{1}\{i_t=i,j_t=j,y_t=1\}
    +\mathbb{1}\{i_t=j,j_t=i,y_t=0\}
    \right]
\end{equation}
denote wins by $i$ over $j$. Grouping trials by the winning and losing items gives the equivalent count likelihood
\begin{equation}
    \Pr(\mathcal D\mid\boldsymbol\theta,\boldsymbol\pi,\boldsymbol\alpha)
    \propto \prod_{i\ne j}\left[
    \sum_{k=1}^{K_{\max}}\pi_k p_{ij}^{(k)}
    \right]^{a_{ij}}.
\end{equation}
where the product is over all ordered pairs of items.

\subsection{Priors and Generative Model}
\label{app:priors}

We use a Bayesian formulation of the model to regularize inference, as BT models and finite mixtures can both face identifiability issues that can be compounded by fusing the two modeling approaches~\cite{newman2023efficient,pearce2025modeling}.

\paragraph{Scores.}
The BT likelihood depends only on score differences $\theta_{ik}-\theta_{jk}$, so the absolute scale of the scores is not identified. We break this symmetry by constraining the scores in each component to sum to zero,
\begin{equation}
    \sum_{i=1}^{N}\theta_{ik}=0,\qquad k=1,\dots,K_{\max},
\end{equation}
and place a zero-mean Gaussian prior on the resulting $(N{-}1)$-dimensional subspace, calibrated so that score \emph{differences} have unit variance:
\begin{equation}
    \bm{\theta}_k \sim \mathcal{N}\!\left(\bm{0},\;\tfrac{1}{2}\!\left(\bm{I}-\tfrac{1}{N}\bm{1}\bm{1}^{\top}\right)\right).
\end{equation}
This implies $\operatorname{Var}(\theta_{ik}) = \tfrac{1}{2}(1-1/N)$ and $\operatorname{Var}(\theta_{ik}-\theta_{jk})=1$, putting the implied prior on pairwise win probabilities $\sigma(\theta_{ik}-\theta_{jk})$ on a sensible scale. Because the prior is proper, a unique posterior mode exists regardless of the connectivity of the empirical comparison graph, avoiding the strong-connectivity requirement of maximum-likelihood BT estimation~\cite{newman2023efficient}. 
For sampling, we reparameterize in $\mathbb{R}^{N-1}$ using any orthonormal basis $Q$ of the sum-to-zero subspace.

\paragraph{Noise-augmentation.}
The prior on the per-component noise parameters is modeled with a symmetric beta prior,
\begin{equation}
    \alpha_k \sim \mathrm{Beta}(1.5,1.5).
\end{equation}

\paragraph{Mixture Components}
Rather than select the number of components directly, we place a Dirichlet Process (DP) prior over the mixing measure with concentration parameter $\gamma$ and truncate it at $K_{\max}$. The base distribution is over the per-component parameters $(\alpha_k, \widetilde{\bm{\theta}}_k)$.
The DP induces the following truncated stick-breaking representation~\cite{Sethuraman1994} of the mixture weights; the finite approximation converges exponentially fast to the true DP~\cite{Ishwaran2001a}:
\begin{equation}
    \begin{cases}
        v_k \sim \mathrm{Beta}(1,\gamma), & k=1,\ldots,K_{\max}-1, \\
        \pi_k = v_k \prod_{h < k}(1 - v_h), & k=1,\ldots,K_{\max}-1, \\
        \pi_{K_{\max}} = \prod_{h=1}^{K_{\max}-1}(1 - v_h).
    \end{cases}
\end{equation}
The mixture assignments are drawn from a categorical distribution defined by $\bm{\pi}$,
\begin{equation}
    Z_t \sim \mathrm{Categorical}(\bm{\pi}).
\end{equation}
We fix $\gamma$ throughout. 
The concentration $\gamma$ has a direct interpretation through the Chinese Restaurant Process representation of the DP: the expected number of occupied components after $m$ comparisons is
approximately $\gamma \log(1 + m/\gamma)$, so $\gamma$ controls the prior expected number of distinct components on a logarithmic scale. 
We choose $\gamma$ to place reasonable prior mass on the range of mixture sizes consistent with our experimental setup.

\subsection{Generative Model}
\label{sec:generative_model}

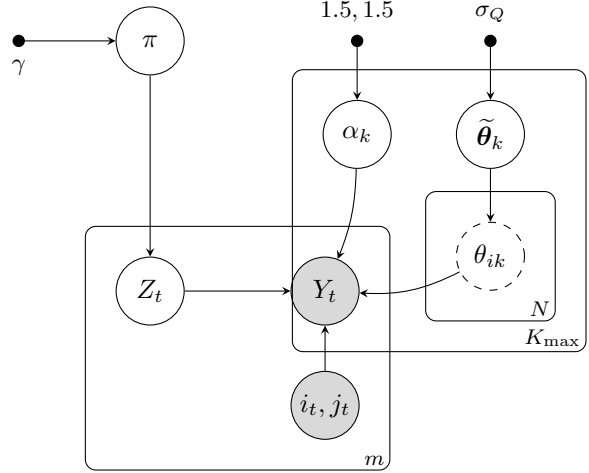
\begin{figure}[htbp]
\centering

\begin{tikzpicture}[
    node distance = 1.0cm and 1.2cm,
    obs/.style    = {circle, draw, fill=gray!30, minimum size=0.9cm, inner sep=1pt},
    latent/.style = {circle, draw, minimum size=0.9cm, inner sep=1pt},
    det/.style    = {circle, draw, dashed, minimum size=0.9cm, inner sep=1pt},
    hyper/.style  = {circle, draw, fill=black, inner sep=1.5pt, minimum size=0pt},
    plate/.style  = {draw, rounded corners, inner sep=0.4cm},
    >=stealth,
]
\node[hyper] (gamma) {};
\node[below=0.05cm of gamma] {\small $\gamma$};
\node[latent, right=of gamma] (pi) {$\pi$};
\node[hyper, right=2.2cm of pi] (ab) {};
\node[above=0.05cm of ab] {\small $1.5,1.5$};
\node[latent, below=0.7cm of ab] (alpha) {$\alpha_k$};
\node[hyper, right=1.6cm of ab] (sigQ) {};
\node[above=0.05cm of sigQ] {\small $\sigma_Q$};
\node[latent, below=0.7cm of sigQ] (theta) {$\widetilde{\boldsymbol{\theta}}_k$};
\node[det, below=0.7cm of theta] (lam) {$\theta_{ik}$};
\node[latent, below=2.4cm of pi] (z) {$Z_t$};
\node[obs, right=1.4cm of z] (y) {$Y_t$};
\node[obs, below=0.6cm of y] (wl) {$i_t,j_t$};
\draw[->] (gamma) -- (pi);
\draw[->] (ab)    -- (alpha);
\draw[->] (sigQ)  -- (theta);
\draw[->] (theta) -- (lam);
\draw[->] (pi)    -- (z);
\draw[->] (z)     -- (y);
\draw[->] (alpha) to[bend left=10]  (y);
\draw[->] (lam)   to[bend left=15]  (y);
\draw[->] (wl)    -- (y);
\node[plate, fit=(lam),
      label={[anchor=south east, inner sep=2pt]south east:{\footnotesize $N$}}] (items) {};
\node[plate, fit=(alpha) (theta) (items),
      label={[anchor=south east, inner sep=2pt]south east:{\footnotesize $K_{\max}$}}] (comp) {};
\node[plate, fit=(z) (y) (wl),
      label={[anchor=south east, inner sep=2pt]south east:{\footnotesize $m$}}] (obsplate) {};
\end{tikzpicture}
\caption{
    Plate diagram for the \MBT\ model.
    Shaded nodes are observed, dashed nodes are deterministic, filled black dots are fixed hyperparameters. 
    The component plate ($k = 1, \ldots, K_{\max}$) contains the per-component luck parameter $\alpha_k$, reduced scores $\widetilde{\boldsymbol{\theta}}_k$, and item strengths $\lambda_{ik}$ (nested over items, $i = 1, \ldots, N$).
    The mixture weights $\pi$ are drawn from a stick-breaking process with concentration $\gamma$ and used to assign each comparison $t = 1, \ldots, m$ to a component $Z_t$, which together with the observed item indices $(i_t,j_t)$ generates the outcome $Y_t$.}
\label{fig:plate-diagram}
\end{figure}

The full generative model can be summarized as follows (see Fig.~\ref{fig:plate-diagram}).
Let $N$ denote the number of items, $m$ the number of pairwise comparisons, and $K_{\max}$ the truncation level for the stick-breaking prior.
For each comparison $t$, let $(i_t,j_t)$ denote the ordered pair shown to the model and let $Y_t$ denote its binary choice.

\smallskip
\noindent\emph{Mixture weights} (stick-breaking, $k = 1, \ldots, K_{\max}-1$):
\begin{equation*}
    v_k \sim \mathrm{Beta}(1,\gamma), \qquad
    \pi_k = v_k \prod_{h<k}(1-v_h) \quad (k<K_{\max}), \qquad
    \pi_{K_{\max}} = \prod_{h=1}^{K_{\max}-1}(1-v_h).
\end{equation*}

\noindent\emph{Per-component parameters} ($k = 1, \ldots, K_{\max}$):
\begin{equation}
    \alpha_k \sim \mathrm{Beta}(1.5,1.5), \qquad
    \widetilde{\boldsymbol{\theta}}_k \sim \mathcal{N}\!\left(\boldsymbol{0}, \sigma_Q^2 \boldsymbol{I}_{N-1}\right), \qquad
    \boldsymbol{\theta}_k = Q\widetilde{\boldsymbol{\theta}}_k, \qquad
    \lambda_{ik} = e^{\theta_{ik}}.
\end{equation}

\noindent\emph{Observations} ($t = 1, \ldots, m$):
\begin{equation}
    Z_t \sim \mathrm{Categorical}(\boldsymbol{\pi}), \qquad
    U_t \mid Z_t = k \sim \mathrm{Bernoulli}(\alpha_k),
\end{equation}
\begin{equation}
    Y_t \mid Z_t = k, U_t \sim
\begin{cases}
    \mathrm{Bernoulli}(1/2) & \text{if } U_t = 1, \\
    \mathrm{Bernoulli}\!\left(\dfrac{\lambda_{i_t k}}{\lambda_{i_t k}+\lambda_{j_t k}}\right) & \text{if } U_t = 0.
\end{cases}
\end{equation}

\section{Inference and Sampling}
\label{app:inference}

We use Hamiltonian Monte Carlo (HMC) to generate samples from the posterior~\cite{duane_hybrid_1987,neal_mcmc_2011}. 
Specifically, we use the No-U-Turn Sampler (NUTS) as implemented in the BlackJAX Python package~\cite{hoffman_no-u-turn_2014,cabezas_blackjax_2024}.
Label switching creates a multimodal posterior for mixture models.
Even though the DP prior technically breaks this symmetry, we observe strong empirical identifiability issues with label switching, and thus use the Hungarian algorithm~\cite{kuhn_hungarian_1955} for post-hoc alignment of chains across sweeps. 

In our estimation procedure, we use a diagonal mass matrix and a target acceptance rate of 0.8. 
Each dataset is fit with 4 chains of 6,000 warmup and 4,000 post-warmup samples (16,000 total draws). 
To reduce label switching during warmup, we run adaptation on a single chain and broadcast the resulting step size and mass matrix to all remaining chains before sampling as recommended in the  \href{https://mc-stan.org/docs/stan-users-guide/problematic-posteriors.html#label-switching-problematic.section}{Stan User Guide}.
Convergence is assessed with the Gelman-Rubin statistic $\hat{R}$ and the effective sample size of both the mixture weights $\bm{\pi}$ and noise parameters $\bm{\alpha}$. 
A fit is deemed acceptable if all components satisfy $\hat{R} < 1.01$, $\text{ESS} \geq 100$, and zero divergent transitions are observed.

\section{Consistency of the estimation model}
\label{app:simulation}
As mentioned before, mixtures and Bradley--Terry models can both face identifiability issues, and our proposal is to combine both.
To verify that the parameters of our implementation can be recovered in practice, we conduct a consistency study in which we start from known ground-truth parameters, generate data with the model, apply the inference algorithm, and validate agreement.
The results below show that the estimation algorithm recovers ground-truth parameters across most of the sweep grid, with two regimes where it fails: 1) when the scores of different components are highly correlated and the number of components is large, and 2) when noise augmentation dominates, regardless of the number of ground-truth components.

\subsection{Data generation grid}
We generate preference matrices from the generative model specified in Section~\ref{sec:generative_model}, with two modifications to explore potential failure modes.

First, we control the correlation between components by drawing each item's score vector $\bm{\theta}_i=(\theta_{i1},\ldots,\theta_{iK_{\mathrm{true}}})$ from $\mathcal{N}(\bm{0},\Sigma(\rho))$, where $\Sigma(\rho)=\sigma^2[(1-\rho)I_{K_{\mathrm{true}}}+\rho\mathbf{1}\mathbf{1}^\top]$.
In many settings, we expect scores across different components to be correlated; e.g., consequentialist and contrarian systems of ethics may disagree on a number of specific dilemmas but will still choose the same option in a majority of scenarios.
Large $\rho$ can model such dependencies and make components harder to tell apart if all other parameters are held constant.

Second, we vary the level of noise by altering the expectation of $\alpha$.
Large values of $\alpha$ decrease the signal-to-noise ratio, making more of the individual pairwise comparisons uninformative. 

We couple these modifications with a sweep over the number of ground-truth components, $K_{\text{true}}$, and hyperparameters of the inference process.

\subsection{Evaluation metrics}
Let $K_{\min}=\min(K_{\mathrm{true}},K_{\max})$. We quantify recovery using three metrics. For mixture weights, we compute the KL divergence between the ground-truth weights and the posterior-mean weight vector:
\begin{equation}
    D_{\mathrm{KL}}\!\left(\bm{\pi}_{\text{gt}} \,\Big\|\, 
    \bar{\bm{\pi}}\right) = \sum_{k=1}^{K_{\min}} \pi_{\text{gt},k} \log
    \frac{\pi_{\text{gt},k}}{\bar{\pi}_k},
\end{equation}
where $\bar{\bm{\pi}} = \tfrac{1}{S}\sum_{s=1}^{S} \bm{\pi}^{(s)}$ is the posterior mean, and both vectors are sorted in descending order before comparison.

For scores, we first match components by descending mixture weights and center the scores in each component.
We then report the Pearson correlation between the ground-truth and posterior-mean scores across all $N \times K_{\min}$ item--component entries.

Finally, for the luck parameters, we compute the mean absolute error between the sorted posterior-mean and ground-truth $\alpha$ values:
\begin{equation}
    \frac{1}{K_{\min}} \sum_{k=1}^{K_{\min}} 
    \left|\bar{\alpha}_k - \alpha_k^{\text{gt}}\right|,
\end{equation}
where $\bar{\alpha}_k = \tfrac{1}{S}\sum_{s=1}^{S} \alpha_k^{(s)}$ and both vectors are sorted in descending order. The use of $K_{\min}$ handles cases where the truncation level differs from the ground truth.

\subsection{Results}
\begin{figure}[H]
    \centering
    \includegraphics[width=\linewidth]{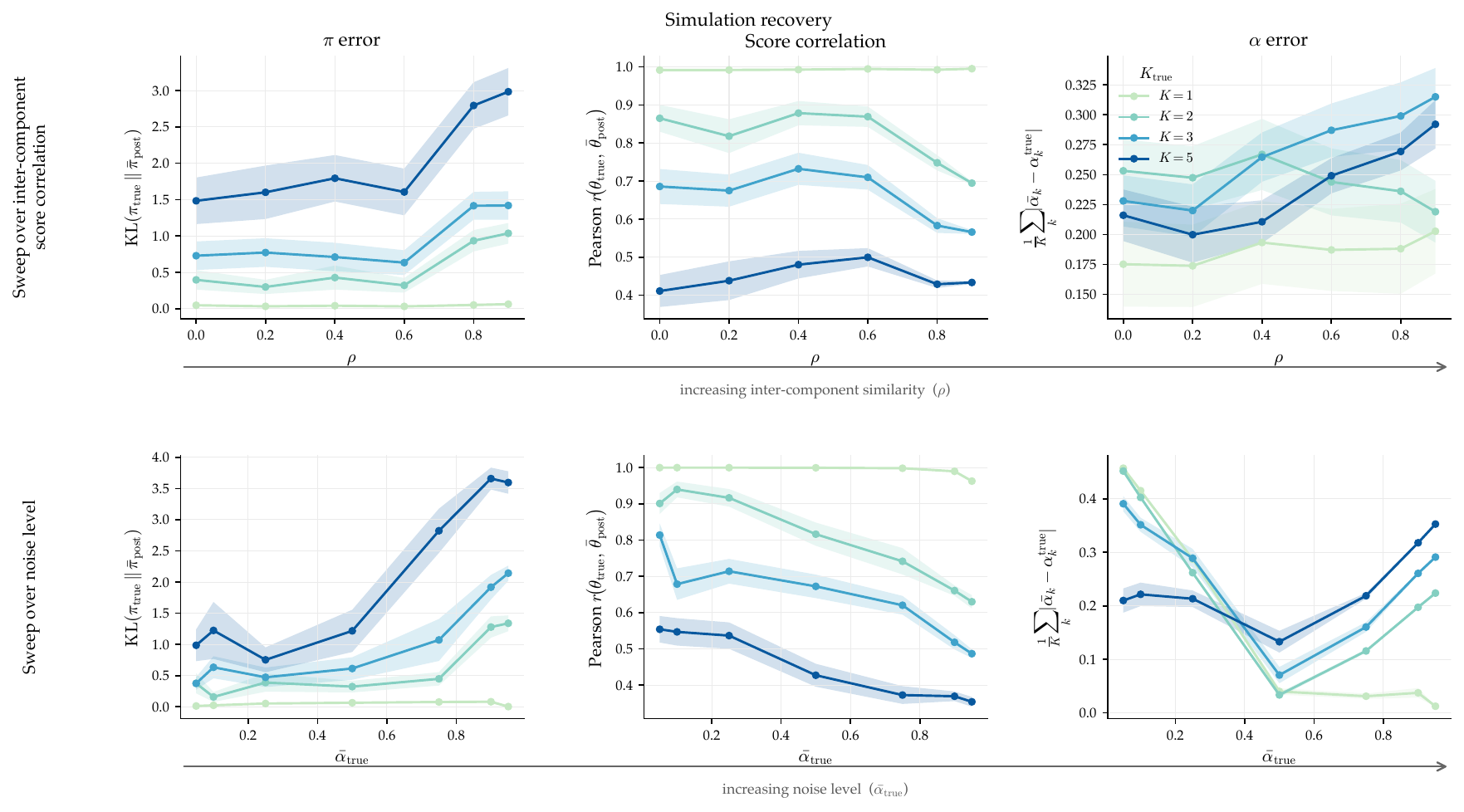}
    \caption{
    \textbf{Top row}: Recovery as a function of inter-score correlation 
    $\rho$, holding the mean luck parameter at $\bar{\alpha} = 0.0$. 
    Mixture-weight KL divergence (left), score Pearson correlation (center), 
    and luck-parameter MAE (right) are plotted for different numbers of 
    ground-truth components $K_{\text{true}}$.
    \textbf{Bottom row}: Recovery as a function of mean luck $\bar{\alpha}$, 
    holding the inter-score correlation at $\rho = 0$.
    }
    \label{fig:sim_recovery_error_sweep}
\end{figure}

Figure~\ref{fig:sim_recovery_error_sweep} plots the results of a recovery sweep.
In the full factorial sweep, the number of mixture components $K_{\text{true}}$ takes values $\{1, 2, 3, 5\}$.
The inter-component score correlation $\rho \in \{0.0, 0.2, 0.4, 0.6, 0.8\}$ controls how geometrically similar the components are to one another.
The mean luck parameter $\bar{\alpha} \in \{0.05, 0.10, 0.20, 0.50, 0.90\}$ controls the tie rate, and therefore how discriminating the pairwise comparisons are.
We also vary the DP concentration used at inference, $\gamma \in \{0.1,0.5\}$, which is a fitting parameter rather than a generative one.
Each cell of the resulting grid is repeated three times with independent synthetic datasets. The remaining settings are fixed.

\begin{figure}[H]
    \centering
    \includegraphics[width=\linewidth]{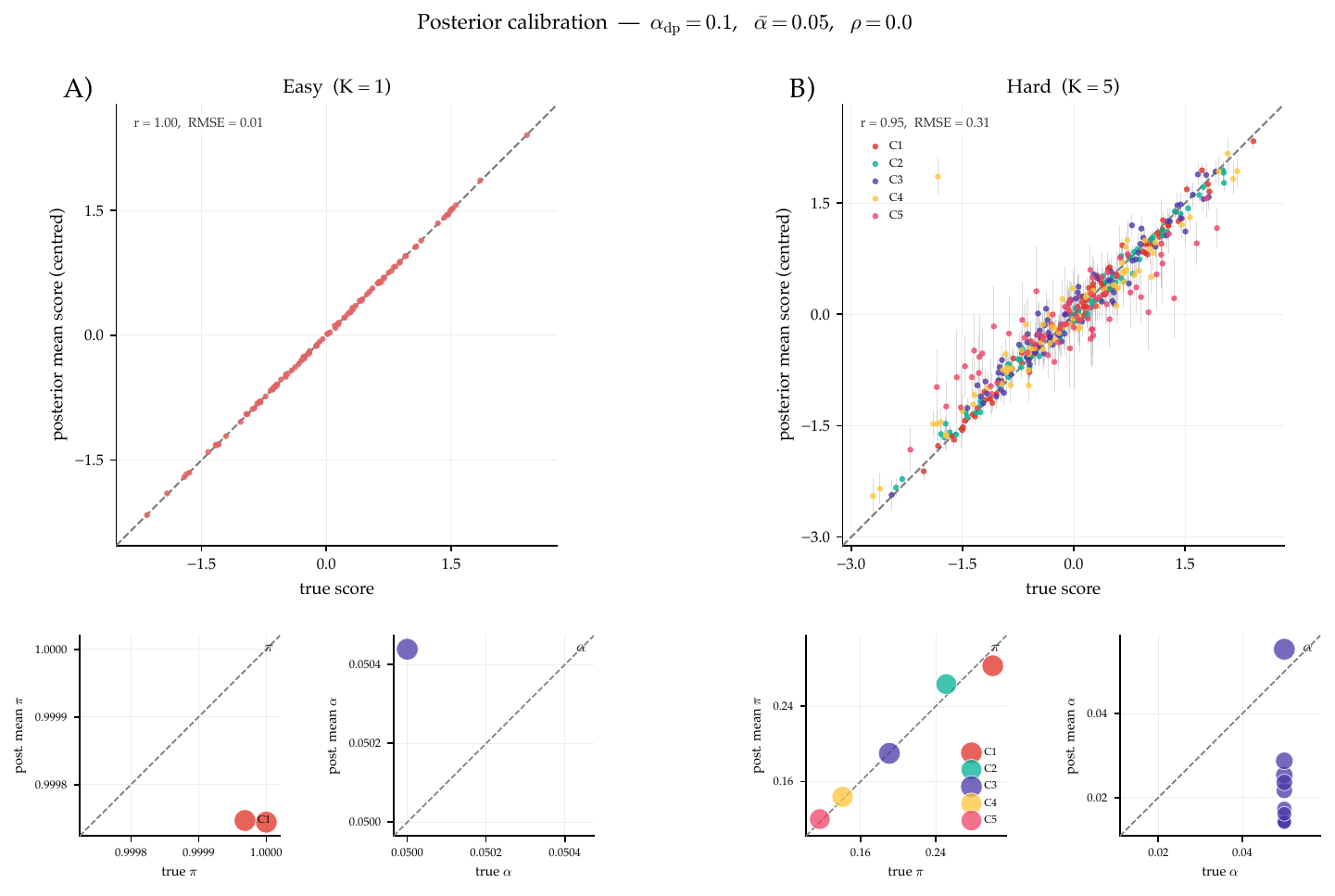}
    \caption{ Posterior calibration of the \MBT\ model. True vs.\ posterior mean scores (main panels), mixture weights $\pi$ (bottom left), and noise scale $\alpha$ (bottom right) for $K_{\mathrm{true}}=1$ (A) and $K_{\mathrm{true}}=5$ (B) ground truths. Point size encodes posterior standard deviation; colours identify mixture
  components.}
\label{fig:sim_recovery_calibration}
\end{figure}

Figure~\ref{fig:sim_recovery_calibration} plots calibration for two examples at $K_{\mathrm{true}}=1$ and $K_{\mathrm{true}}=5$. We compare the true and posterior mean scores, mixture weights, and $\alpha$ values. At $K_{\mathrm{true}}=1$, we achieve an almost perfect reconstruction; at $K_{\mathrm{true}}=5$, we reconstruct the order but not necessarily the exact score values.

Score recovery is near-perfect under $K_{\mathrm{true}}=1$ ($r=1.00$, RMSE$=0.01$), confirming the model is not
  over-parameterized on simple data. Under $K_{\mathrm{true}}=5$ recovery degrades modestly ($r=0.95$, RMSE$=0.31$), with
  the bulk of residual error concentrated in the minor components (C4–C5): their scores scatter further
  from the diagonal and carry wider error bars, reflecting the smaller effective sample size available to
  identify low-weight components. The dominant component (C1) remains well-recovered even at $K_{\mathrm{true}}=5$. Mixture
   weights are recovered accurately across all five components, with posterior means tracking the true
  $\pi$ values along the diagonal and uncertainty (point size) increasing for minor components. The noise
  scale $\alpha$ is consistently underestimated under $K_{\mathrm{true}}=5$. These
  results establish that the DP inference is well-calibrated for the dominant structure while accurately
  flagging elevated uncertainty for minor components.

\section{MBT inference on LLM Preference Data}
\label{app:mbt_llm}

This appendix provides additional qualitative views of the \MBT\ fits discussed in Section~\ref{sec:mixture_results}. These figures are not used as separate evidence for new claims. They document where the fitted components differ from the aggregate ordering and help interpret the main results.

\subsection{\Population}
\label{app:population_data}

\begin{figure}[H]
    \centering
    \includegraphics[width=\linewidth, trim=0cm 2cm 0cm 1cm, clip=true]{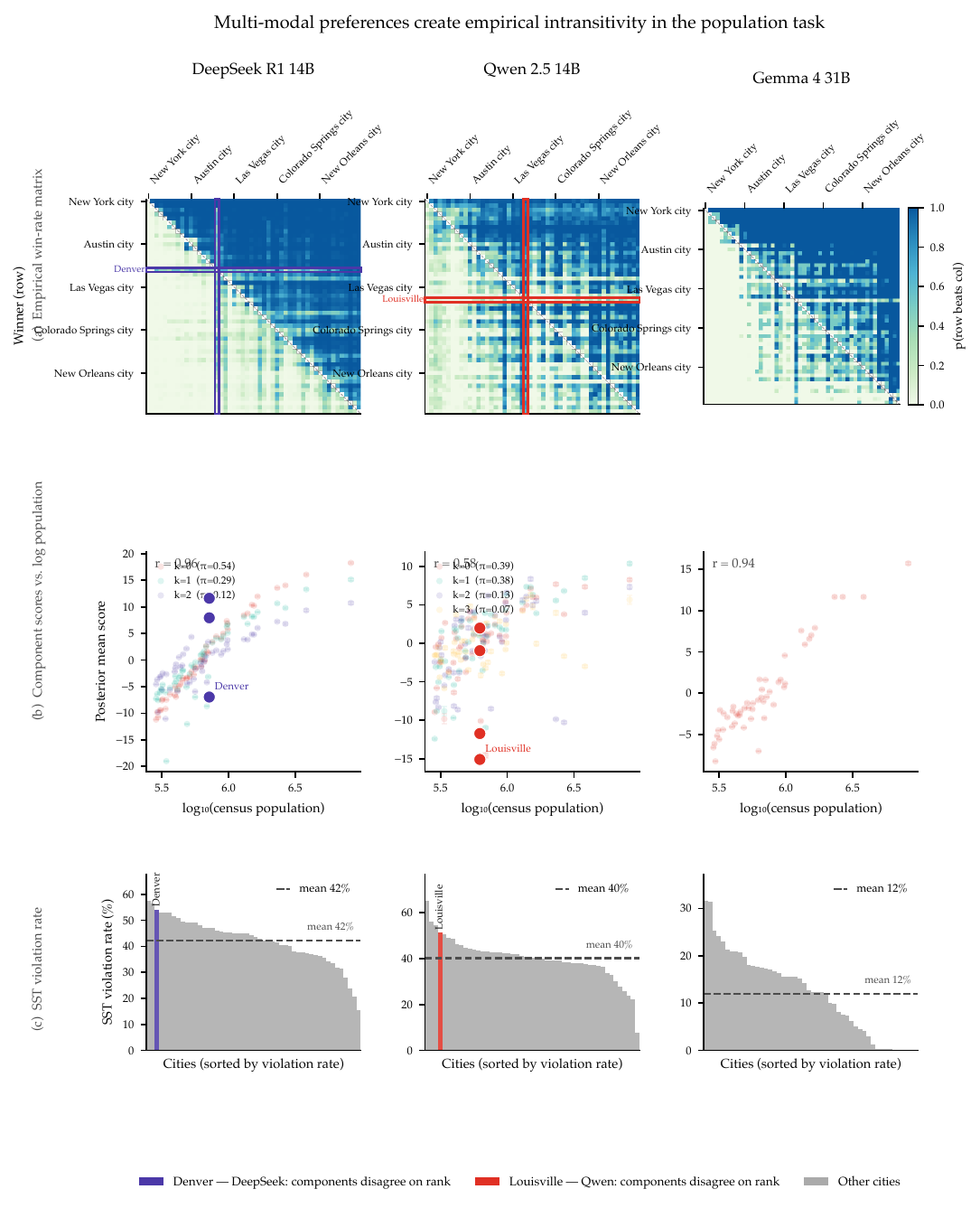}
    \caption{
    \MBT\ fits on the \Population task for \DeepSeek, \QwenSmall, and \GemmaFour.
    \textbf{Top:} Empirical win-rate matrices $p_{ij}$, with cities sorted by true Census population from largest to smallest. Denver and Louisville are highlighted as examples of cities whose aggregate ranking is unstable across components.
    \textbf{Middle:} Posterior mean Bradley--Terry scores plotted against log Census population. Points are colored by mixture component, with Denver and Louisville highlighted across components.
    \textbf{Bottom:} Per-city SST violation rates, sorted from largest to smallest. The highest-violation cities correspond to items whose inferred component scores disagree most strongly.
    }
    \label{fig:population_intransitivity}
\end{figure}

We test each model's preferences on a 50-city pairwise-comparison task, ranking U.S. cities by population. In the empirical win-rate matrices in Figure~\ref{fig:population_intransitivity}, cities are ordered along both axes by their true Census population. A perfectly transitive ranker would produce a smooth gradient away from the diagonal, with steeper gradients indicating more deterministic preferences. Deviations from this pattern indicate pairwise reversals that are not explained by the true population order.

We fit the \MBT\ model with $K_{\max}=5$ to each model's comparisons. The fitted models differ sharply in effective component count: \DeepSeek ($k_{\mathrm{eff}} \approx 3$) and \QwenSmall ($k_{\mathrm{eff}} \approx 4$) split into multiple modes, while \GemmaFour ($k_{\mathrm{eff}} \approx 1$) is nearly unimodal. Posterior mean Bradley--Terry scores correlate strongly with log population for \DeepSeek and \GemmaFour, but weakly for \QwenSmall.

Intransitivity is not uniform across cities. Louisville and Denver, both midsize cities whose ranking is contested across latent components, consistently sit near the top of the SST violation distribution. Here SST requires that if $p_{ij} \geq 0.5$ and $p_{jk} \geq 0.5$, then $p_{ik} \geq \max(p_{ij}, p_{jk})$. For \DeepSeek, Denver's component scores span roughly 25 units: one component ranks it near the top, while another ranks it near the bottom. Louisville shows the same pattern under \QwenSmall. Aggregate SST violation rates reflect this heterogeneity: 42\% for \DeepSeek and 40\% for \QwenSmall, compared with 12\% for \GemmaFour.

\subsection{\MoralMachines}
\label{app:mbt_moral}

\begin{figure}[H]
    \centering
    \includegraphics[width=0.58\linewidth]{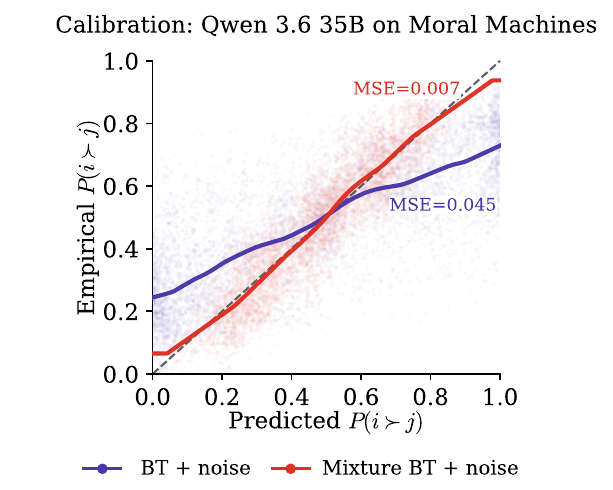}
    \caption{
    Posterior predictive calibration for \QwenLarge on \MoralMachines.
    Each point is an item pair. The $x$-axis shows the model-predicted probability that item $i$ is preferred to item $j$, and the $y$-axis shows the empirical win rate. The \MBT\ model lies closer to the diagonal than the single-component noise-augmented Bradley--Terry baseline, indicating better calibration of pairwise probabilities.
    }
    \label{fig:mm_pp_calibration}
\end{figure}

\begin{figure}[H]
    \centering
    \includegraphics[width=0.72\linewidth]{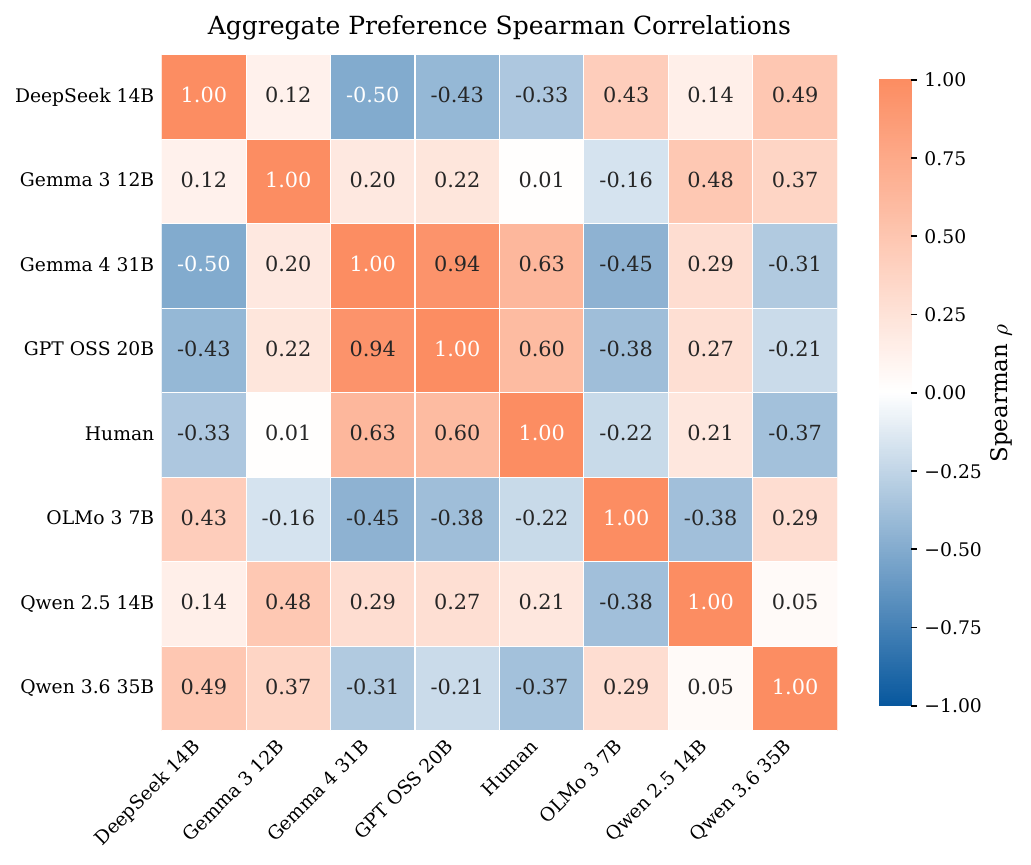}
    \caption{
    Aggregate preference similarity on \MoralMachines.
    Each cell reports the Spearman correlation between two models' aggregate preference rankings. The heatmap gives the model-level summary that is usually reported by single-ranking analyses.
    }
    \label{fig:mm_aggregate_heatmap}
\end{figure}

\begin{figure}[H]
    \centering
    \includegraphics[width=0.9\linewidth]{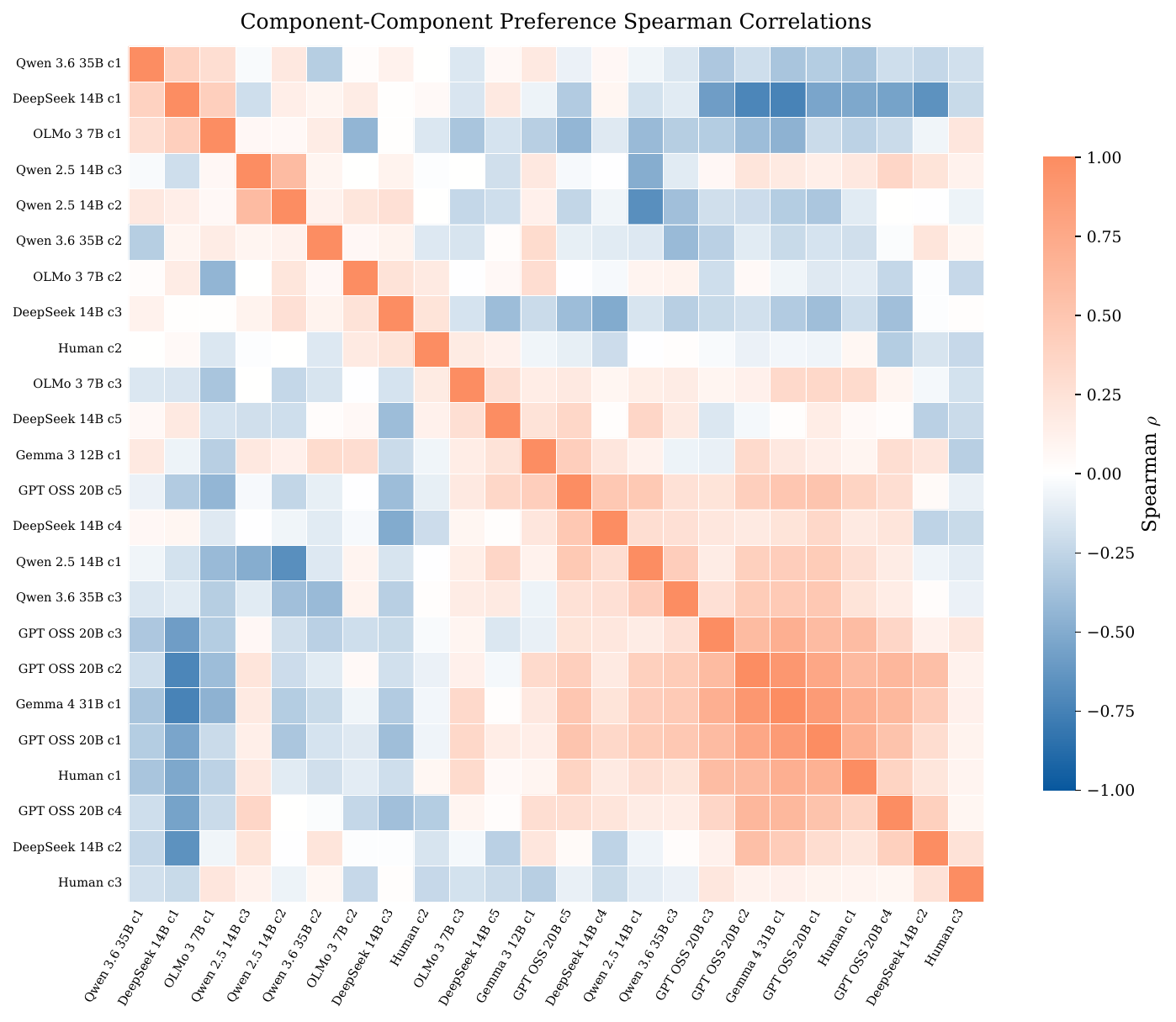}
    \caption{
    Component-level preference similarity on \MoralMachines.
    Each cell reports the Spearman correlation between a pair of inferred mixture components, including human components where available. Blocks of positive and negative correlation reveal structure that is compressed away in the aggregate heatmap.
    }
    \label{fig:mm_component_heatmap}
\end{figure}

Section~\ref{sec:hetero} argues that aggregate model similarity can hide heterogeneous component-level structure. The \MoralMachines\ task is the clearest case because all models compare the same moral dilemmas, but their mixture components do not align in a one-to-one way. Figures~\ref{fig:mm_pp_calibration}--\ref{fig:mm_component_heatmap} provide the supporting views used for that discussion.

Figure~\ref{fig:mm_pp_calibration} compares empirical pairwise win rates with posterior predictive probabilities for \QwenLarge on \MoralMachines. The \MBT\ model tracks the empirical probabilities more closely than the single-component noise-augmented Bradley--Terry baseline, consistent with the held-out predictive gains reported in the main text.

The main aggregation analysis compares model-level rankings with component-level rankings. Figure~\ref{fig:mm_aggregate_heatmap} shows the aggregate Spearman correlations between \MoralMachines preference rankings. Figure~\ref{fig:mm_component_heatmap} shows the corresponding component--component correlations. Read together, the heatmaps show why aggregate agreement is not enough: two models can have similar aggregate rankings while containing component pairs that disagree, and models with weak aggregate agreement can still share aligned latent components.

\section{Compute resources}
\label{app:compute}

We report the compute used for the experiments in the paper. Estimates combine recorded LLM
elicitation metadata, recorded wall-clock times for the transitivity-control jobs, and the SLURM
resource tiers used in the posterior-fitting and validation sweeps. Wall-clock logs for all HMC posterior fits are not available exactly, so CPU-hour totals for those experiments are reported as conservative allocated CPU-hour upper bounds rather than measured consumption.

\paragraph{Hardware environment.}
LLM elicitation was run on an internal GPU cluster using NVIDIA Hopper-class GPUs, with
1--2 GPUs per run, 8 CPU cores, and approximately 80--120 GB RAM per job. Statistical diagnostics, posterior inference, cross-validation, and synthetic recovery experiments were run on a CPU
partition. CPU jobs used 1--4 CPU cores depending on the experiment and
approximately 8--32 GB RAM. Figure extraction and plotting compute was negligible.

\begin{table}[t]
\centering
\small
\setlength{\tabcolsep}{3pt}
\renewcommand{\arraystretch}{1.12}
\caption{
Compute used by result family. GPU-hour totals for LLM elicitation are measured from run
metadata. CPU-hour totals for MCMC-based posterior fitting, cross-validation, and synthetic
recovery are conservative allocated upper bounds because complete wall-clock logs were not
available for all runs.
}
\label{tab:compute-resources}
\begin{tabularx}{\linewidth}{@{}p{0.22\linewidth}p{0.18\linewidth}Xr@{}}
\toprule
Result family & Runs & Resources per run & Total estimate \\
\midrule
LLM elicitation
& 28 model--task runs
& 1--2 Hopper GPUs; 8 CPU; 80--120 GB RAM
& $\approx 251$ GPU-h \\

Transitivity diagnostics
& 243 jobs
& 1 CPU core; modest RAM
& $\approx 1.1$ CPU-h \\

Main MBT posterior fits
& 25 fits
& 4 CPU; 16--32 GB RAM; 8--24 h allocation
& 800--2,400 CPU-h \\

5-fold CV, MBT
& 140 fits
& 4 CPU; 16--32 GB RAM; 8--24 h allocation
& 4,480--13,440 CPU-h \\

5-fold CV, $K_{\max}=1$ baseline
& 140 fits
& 2 CPU; 8--16 GB RAM; 4--12 h allocation
& 1,120--3,360 CPU-h \\

Synthetic recovery
& 720 fits
& 4 CPU; 16--32 GB RAM; 6--18 h allocation
& 17,280--51,840 CPU-h \\

\Population top-50 appendix
& 6 fits
& 4 CPU; 16--32 GB RAM; 8--24 h allocation
& 192--576 CPU-h \\

Figure extraction and plotting
& cache extraction and plotting
& Cluster CPU and local CPU
& small \\
\bottomrule
\end{tabularx}
\end{table}

\paragraph{LLM elicitation.}
Table~\ref{tab:llm-elicitation-compute} reports the final LLM elicitation runs used in the paper.
These runs cover seven models and four datasets. The total compute for the final reported elicitation
runs was approximately 251 GPU-hours. Including reruns, preliminary runs, failed or discarded
runs, and earlier debugging runs, the recorded elicitation total was approximately 1,155 GPU-hours.

\begin{table}[h]
\centering
\small
\setlength{\tabcolsep}{5pt}
\caption{
Final LLM elicitation compute by model. Wall time is aggregated across the four reported dataset
runs for each model. GPU-hours are computed as wall time multiplied by the number of GPUs
allocated per run.
}
\label{tab:llm-elicitation-compute}
\begin{tabular}{lrrr}
\toprule
Model & GPUs/run & Wall time (h) & GPU-hours \\
\midrule
\QwenLarge & 2 & 88.9 & 177.8 \\
\DeepSeek & 1 & 44.4 & 44.4 \\
\GptOss & 2 & 10.8 & 21.5 \\
\GemmaFour & 2 & 1.6 & 3.1 \\
\GemmaThree & 2 & 0.8 & 1.6 \\
\OLMo & 2 & 0.7 & 1.4 \\
\QwenSmall & 2 & 0.5 & 0.9 \\
\midrule
Total & -- & -- & 250.7 \\
\bottomrule
\end{tabular}
\end{table}

\paragraph{Posterior inference.}
The main \MBT\ posterior fits used four chains with NUTS sampling. Each final
\MBT\ fit was allocated 4 CPU cores and 16--32 GB RAM for 8--24 hours.
Across 25 final empirical fits, this corresponds to 800--2,400 allocated CPU-hours. These are upper
bounds because jobs often finished before the allocation limit.

\paragraph{Cross-validation.}
Held-out prediction experiments used five folds for each model--dataset pair and compared the
\MBT\ model against single-component baselines. The \MBT\ cross-validation
sweep used 140 fits allocated 4 CPU cores for 8--24 hours each, giving 4,480--13,440 allocated
CPU-hours. The $K_{\max}=1$ baseline sweep used 140 fits allocated 2 CPU cores for 4--12 hours each,
giving 1,120--3,360 allocated CPU-hours.

\paragraph{Synthetic recovery.}
The synthetic recovery study was the largest CPU component. It used 720 posterior fits over a grid
of ground-truth component counts, inter-component score correlations, noise levels, DP concentration
settings, and random seeds. Each fit was allocated 4 CPU cores and 16--32 GB RAM for 6--18 hours,
for an allocated upper bound of 17,280--51,840 CPU-hours. This sweep validates the inference
procedure but is not required to reproduce the main empirical results.

\paragraph{Preliminary and unreported compute.}
The full research project used more compute than the final reported experiments. In particular, the recorded LLM elicitation total was approximately 1,155 GPU-hours when including reruns,
preliminary runs, failed parsing runs, and debugging runs, compared with approximately 251
GPU-hours for the final reported elicitation runs. Additional CPU time was used for exploratory
posterior fits, plotting, cache extraction, and diagnostic scripts; these costs were small compared with
the posterior inference and synthetic recovery sweeps.

\end{document}